\documentclass[runningheads]{llncs}

\usepackage{graphicx}
\usepackage{comment}
\usepackage{amsmath,amssymb}
\usepackage{color}
\usepackage{url}
\usepackage{hyperref}

\usepackage{xcolor, colortbl}
\usepackage{xspace}
\usepackage{enumitem}
\usepackage[capitalize,noabbrev]{cleveref}
\usepackage{pgfplots}
\usepackage{filecontents}
\usepackage{caption}
\usepackage{graphicx}
\usepackage{wrapfig}
\usepackage{ragged2e}  
\usepackage{cite}
\usepackage{lmodern}
\usepackage{doi}
\usepackage{subcaption}

\pgfplotsset{compat=1.17}

\colorlet{dgreen}{green!50!black}
\colorlet{dred}{red!50!black}
\colorlet{dblue}{blue!70!black}

\newcommand{\seq}[1]{({\fontfamily{lmtt}\selectfont{#1}})}

\definecolor{darkbrown}{rgb}{0.4, 0.26, 0.13} 

\newif\ifreview
\reviewfalse

\ifreview
	\usepackage{lineno}

	\linenumbers
\fi

\begin{document}


\def\SubNumber{068}

\def\GCPRTrack{Main Track}

\title{Unlocking In-Context Learning for Natural Datasets Beyond Language Modelling}

\ifreview
\else

	\author{Jelena Bratuli\'{c}\inst{1}\thanks{Corresponding author. Email: \email{bratulic@cs.uni-freiburg.de}} \and
	Sudhanshu Mittal\inst{1} \and
	David T.~Hoffmann\inst{1} \and \\ Samuel B\"{o}hm\inst{2} \and
	Robin Tibor Schirrmeister\inst{3} \and
	Tonio Ball\inst{2} \and  \\ Christian Rupprecht\inst{4} \and
	Thomas Brox\inst{1}}
	
	\authorrunning{J.Bratuli\'{c} et al.}
	
	\institute{Computer Vision Group, University of Freiburg, \\
        \and Neuromedical A.I. Lab, Medical Center -- University of Freiburg\\
        \and Medical Physics, Medical Center -- University of Freiburg \\
        \and Visual Geometry Group, University of Oxford}
\fi

\maketitle              

\begin{abstract}
Large Language Models (LLMs) exhibit In-Context Learning (ICL), which enables the model to perform new tasks conditioning only on the examples provided in the context without updating the model's weights. While ICL offers fast adaptation across natural language tasks and domains, its emergence is less straightforward for modalities beyond text.  
In this work, we systematically uncover properties present in LLMs that support the emergence of ICL for autoregressive models and various modalities by promoting the learning of the needed mechanisms for ICL.
We identify exact token repetitions in the training data sequences as an important factor for ICL.
Such repetitions further improve stability and reduce transiency in ICL performance. Moreover, we emphasise the significance of training task difficulty for the emergence of ICL.
Finally, by applying our novel insights on ICL emergence, we unlock ICL capabilities for various visual datasets and a more challenging EEG classification task. Code is available at \url{https://github.com/jelenab98/unlocking_icl}

\end{abstract}
\section{Introduction}\label{sec:introduction}

In-context learning (ICL) is a notable emerging feature observed primarily in transformer models, such as Large Language Models (LLMs)~\cite{NEURIPS2020_1457c0d6, radford2019language}. ICL presents the ability to gather information to solve tasks that were not seen during training, such as looking up class labels or learning an algorithm (mapping rule), by solely conditioning on the examples provided in the context. To achieve this, no weight updates or fine-tuning is required; instead, examples are used to define a task within the context during inference. ICL contrasts with the ``classical" in-weight learning (IWL), where the knowledge required for inference tasks is embedded within the model weights during training.
The performance and generalization of the IWL depend on the pretraining task, and it is less flexible, as it does not allow for fast adaptation to new tasks without weight updates via gradient descent.

In-context learning was first described as few-shot learning in LLMs ~\cite{NEURIPS2020_1457c0d6, radford2019language}. Given its rapid adaptation capabilities, it has become a standard way for humans to interact with language models for everyday use and has found various applications in different domains~\cite{agarwal2024manyshot, long-etal-2023-adapt, min-etal-2022-metaicl, pawelczyk2024icl_unlearning, ram-etal-2023-context}. Furthermore, ICL has been proven helpful for vision-language models (VLMs)~\cite{alayrac2022flamingo, laur2023idefics, Emu2}, and even for tabular data~\cite{hollmann2023tabpfn, hollmann2025tabpfn} as it 
enables fast online adaptation, online algorithm learning, adaptation to novel datasets and novel label mappings ~\cite{akyurek2024incontext, NEURIPS2023_b2e63e36, NEURIPS2022_c529dba0, hollmann2025tabpfn}. 
Overall, ICL promises to be a fast and reliable method for new tasks with limited training data.
For instance, applications that require few-shot adaptation to novel users or novel sensor/input data sets, like EEG-based brain-computer interfaces, could greatly benefit from high-quality adaptation methods that do not rely on retraining or fine-tuning the model.

Despite its promising capabilities, the emergence of ICL within models is non-trivial; it only emerges under specific training conditions. 
For instance, training on natural language often elicits strong ICL performance. 
Chan et al.~\cite{NEURIPS2022_77c6ccac} attribute this to particular data distributional properties inherent to natural language, namely 1) burstiness: an increased likelihood to observe a token again, after it was seen recently, 
and 2) skewness: a sharply declining distribution over token frequencies with a long tail data distribution.
Chan et al. further demonstrate the effectiveness of these training properties on Omniglot~\cite{omniglot}, resulting in the emergence of ICL. 
However, as we show here, this does not generalize to more complex vision datasets like
CIFAR~\cite{cifarfs}, Caltech-101~\cite{caltech} and DTD~\cite{dtd}, nor does it transfer to other modalities such as EEG. 
Thus, we ask ourselves: \textbf{What do we need to unlock ICL for more general and arguably noisy datasets and modalities?} 

\begin{figure}[t]
    \centering
    \includegraphics[width=\linewidth]{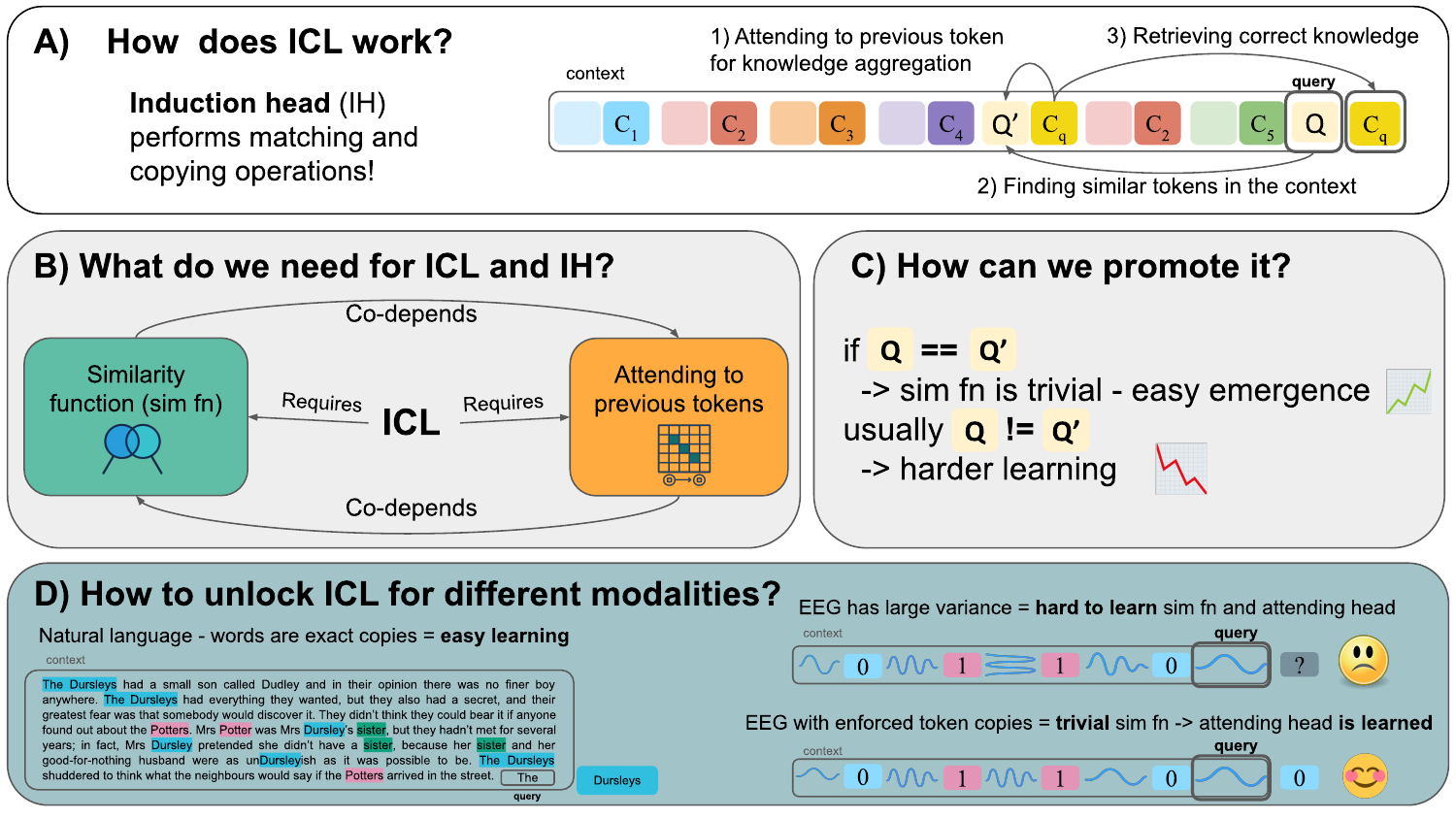}
    \caption{
    A) ICL requires two operations: a similarity function and a head that attends to the previous token for knowledge aggregation; together, they present an induction head. B) A similarity function needs to be established for the previous-token heads to form. Still, the similarity function has no purpose if it can not be associated with relevant knowledge. C) The formation of a previous-token head should be promoted by simplifying the similarity function -- by including exact token copies in the sequence. D) Enforcing exact copies in the sequences enables ICL for noisy and complex data beyond text, such as images and EEG.
    }
    \label{fig:teaser_new}
\end{figure}

Answering this question requires a deeper understanding of ICL; specifically, we need to understand what a model needs to learn for ICL. 
In general, ICL requires: 1) a knowledge aggregation function, which extracts algorithms, rules or information from the context and aggregates this knowledge in specific tokens of the context and 
2) a look-up mechanism that allows retrieving this aggregated information that is relevant to the current last token in a sequence (the so-called query)~\cite{olsson2022context, Reddy2024the, singh2024needs}.
Different ICL tasks will have slight differences in these two functions. In our classification setup, where the sequence contains paired signal-label tokens, the aggregation function gathers information from the previous signal token into the corresponding label token, forming a previous-token head, and the lookup mechanism is a simple similarity function that identifies similar tokens, relevant to the query signal.

Learning from a previous token-attending head does not contribute to ICL unless the similarity (look-up) mechanism is also learned, as there is no learning signal from the loss.
Conversely, learning the look-up function (similarity function) between query and similar tokens is not helpful unless useful information has already been aggregated in those tokens. 
In essence, the learning of each component is interdependent: the similarity function requires that the previous token head has already been learned and learning the previous token head requires the similarity function to be learned (see \cref{fig:teaser_new}B).

These mechanistic insights compel us to investigate why ICL succeeds on language tasks, and Omniglot~\cite{NEURIPS2022_77c6ccac} but fails to generalize effectively to broader datasets and domains.
We believe that the answers lie in the learning interplay of the two components: 1) We show in~\cref{app:burstiness} that language naturally contains many exact copies of tokens and $n$-grams as well as synonyms in a continuous sequence of tokens.
Moreover, prior work has shown that synonyms tend to be clustered or represented closely~\cite{clark-etal-2019-bert, elhelo2025inferringfunctionalityattentionheads,lindsey2025biology, wordvec_neurips, pennington-etal-2014-glove, serina_synonyms, Thieen2023ProbingLL}. We hypothesise that this simplifies the learning of the similarity function, as the required function is close to the identity and, by doing so, it breaks the interdependence between the aggregation and similarity components necessary for ICL to emerge. 
Thus, we argue that introducing exact token repetitions into training sequences -- when they are not naturally present -- can facilitate the learning of ICL.
We further suggest that 2) the relative difficulty (and expected accuracy) of the ICL and the IWL solution influence whether the model prioritizes ICL or not. 
When the IWL task is overly simple, the model may exhibit a simplicity bias, prioritising IWL learning and bypassing in-context learning. We suspect this phenomenon extends to LLMs as well, where language modelling serves as a fairly complex IWL task, thereby encouraging the emergence of ICL. 

In this work, we examine the details of learning ICL in depth, investigating the circumstances under which ICL emerges. 
1) We find that using exact copies of tokens during training facilitates ICL learning and leads to higher ICL accuracy. 
2) We further show that against prior believes \cite{NEURIPS2022_77c6ccac} burstiness is not essential for ICL. A single exact token copy in the context can be sufficient.
3) We present evidence that exact token copies simplify the ICL learning task by reducing the complexity of the similarity function to be learned, giving an initial boost to the ICL learning mechanisms.
4) We further show that ICL vs.~IWL task difficulty is a significant driver of ICL emergence, i.e.~if, the IWL task is difficult and complex, the model is more likely to learn ICL.
5) Finally, we demonstrate that our novel insight unlocks ICL for multiple standard vision datasets and even enables ICL for noisy continuous data, such as EEG, where ICL allows few-shot transfer to novel datasets.

\section{Related work}\label{sec:related_work}

\textbf{In-context learning (ICL).} In-context learning, initially observed as an emerging ability in LLMs~\cite{NEURIPS2022_c529dba0}, enables fast adaptation to various new tasks without gradient updates~\cite{alayrac2022flamingo, bai2023sequentialmodelingenablesscalable, visprompt, hollmann2025tabpfn, NEURIPS2023_398ae57e, zhu2024incoroincontextlearningrobotics}. Plenty of research has been dedicated to understanding how to obtain the best ICL performance by analyzing the importance of pretraining data~\cite{NEURIPS2022_77c6ccac, gu2023pretraininglearncontext, han-etal-2023-understanding, levine2022pretrainig, liu-etal-2022-makes, min-etal-2022-metaicl, wies2023learnability}, demonstration selection and prompt design~\cite{rubin-etal-2022-learning, Suo2024VisualPS, NEURIPS2023_398ae57e, Voronov2024-iv, yang2023auto} or framing ICL as in-context vectors~\cite{liu2024incontextvectorsmakingcontext, huang2024multimodaltaskvectorsenable, peng2024livelearnableincontextvector}. On the other hand, some works~\cite{akyurek2023what, dai-etal-2023-gpt, 52580} provided insights into the ICL working mechanisms by studying ICL on a simple regression task, showing how transformers act as meta-optimisers performing gradient descent.

Numerous works indicate that the training data distribution plays a role in the emergence of ICL, where challenging examples and long-tail tokens, and a large number of rarely occurring classes have been demonstrated to promote ICL~\cite{han-etal-2023-understanding, NEURIPS2022_77c6ccac}, while Razeghi et al.~\cite{razeghi-etal-2022-impact} found a correlation between the input data term frequency and the ICL performance. Furthermore, Chan et al.~\cite{NEURIPS2022_77c6ccac} demonstrate how certain data distributional properties, such as skewed token distribution and burstiness, benefit the ICL in a small synthetic scenario, while Singh et al.~\cite{singh2023the} subsequently showed that the ICL in this setup can become transient, highlighting the conflict between the ICL and IWL circuits. Similarly, Chen et al.~\cite{chen2024parallel} argued that parallel structures, which follow similar semantic or syntactic templates in the pretraining textual data facilitate ICL in language models. Our work builds upon previous studies on the importance of data distributional properties~\cite{NEURIPS2022_77c6ccac, singh2023the} and provides additional insights into unlocking ICL for various modalities and complex data. Concurrent with our work, Zucchet et al.~\cite{zucchet2025emergencesparseattentionimpact} propose a theoretically grounded framework for sparse attention emergence and find that repetitions in data accelerate induction‑head formation and in‑context learning, which supports our finding that exact token repetitions in the training sequences promote in-context learning.

\textbf{Understanding ICL mechanisms.} Mechanistic studies on the emergence of ICL have identified a specialised attention pattern that conducts matching and copying operations as a key mechanism for ICL -- an induction head~\cite{olsson2022context}.
Recent works have been studying the formation of the induction heads and their role for ICL in a simplistic scenario~\cite{Reddy2024the, singh2024needs, NEURIPS2024_icl_markov},  where Reddy~\cite{Reddy2024the} demonstrates with a simple two-parameter model that ICL is driven by the formation of an induction head, which emerges due to nested non-linearities in a multi-layer attention network. 
Our work builds on this interpretability framework to trace the dynamics of the induction heads during training. We explain how certain data distributional properties influence the formation of induction heads and the performance of ICL.

\textbf{Generalisation in EEG for motor imagery.}
Due to individual variability, cross-dataset generalisation in EEG-based motor imagery (MI), although highly desirable,  remains a challenge. While zero-shot EEG methods are increasing, they typically perform multi-modal alignment with EEG and enable classification to unseen classes from the same datasets only~\cite{li5177120two, song2023decoding, liu2302brainclip}. For MI-decoding, pre-trained EEG transformer models show promise but lack zero-shot capabilities \cite{jiang2024large, patil2024coordconformer}. To our knowledge, only \cite{duan2020zero} have explored zero-shot learning for MI-EEG using outlier detection for base and novel classes.
Our work demonstrates how enabling ICL for EEG provides a promising new direction for cross-dataset EEG generalisation without any fine-tuning. 

\section{Experimental setup}\label{sec:set_up}

We investigate how in-context learning (ICL) emerges by training a causal GPT-2 model~\cite{radford2019language} on sequences of image-label pairs from standard few-shot learning datasets: Omniglot~\cite{omniglot}, CIFAR-100~\cite{cifarfs}, Caltech-101~\cite{caltech}, and DTD~\cite{dtd}.

\begin{figure}[bp]
    \centering
    \includegraphics[width=.8\linewidth]{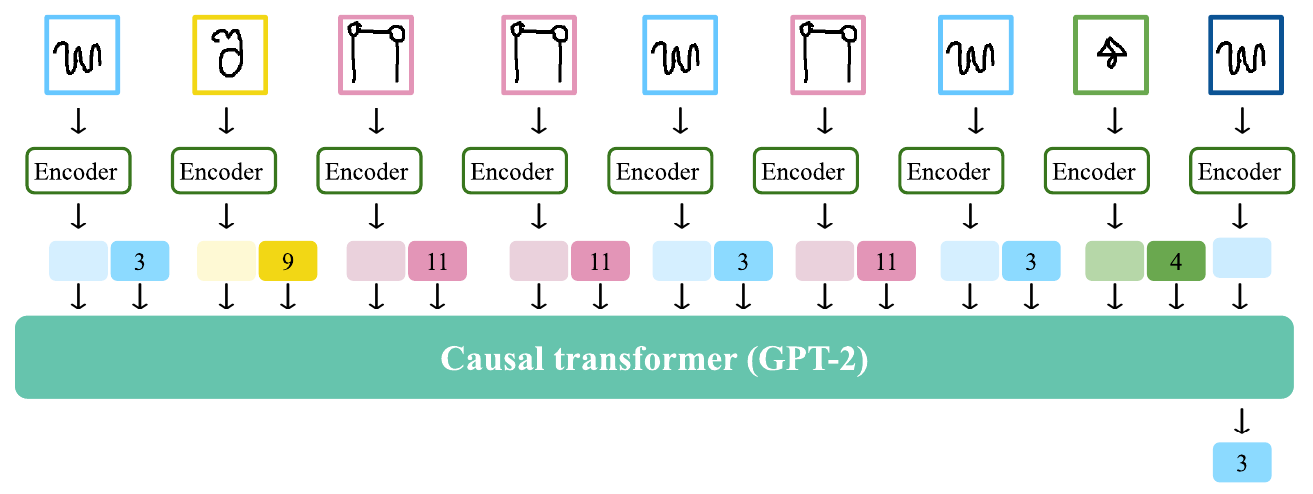}
    \caption{We train GPT-2 as a next-token prediction from scratch with image-label pairs forming a sequence with control of the training sequence distribution.}
    \label{fig:gpt_overview}
\end{figure}

The autoregressive model in this work is trained with a sequence length of $2L+1$ with  $L$ image-label pairs in the context followed by a query image, as shown in Figure~\ref{fig:gpt_overview}. The in-weight learning objective is to predict the label of the last image, which is the $(2L+1)$-th token, given a sequence of $L$ interleaved image-label pairs. Each image-label pair is converted into token embeddings separately. The model is trained to maximize the likelihood of the next token, with the loss applied to the final query output, thus using last-token prediction as the IWL training objective.

\textbf{Training sequences.} We employ a mixture of (1) \textbf{standard sequences}, in which sample-label pairs are uniform randomly selected from the training dataset without any repetitions in the sequence, and (2) \textbf{in-context (bursty) sequences}, where the query image-label information is enforced to be present in the sequence by using a pair similar to the query image-label pair. Using in-context sequences, the model can solve the task without relying solely on the model weights.
The proportion of each sequence type in the total amount of training sequences is treated as a hyper-parameter. Following the setup from~\cite{NEURIPS2022_77c6ccac} we use 10\% of standard and 90\% of in-context sequences.

\begin{figure}[tp]
    \centering
    \includegraphics[width=.8\linewidth]{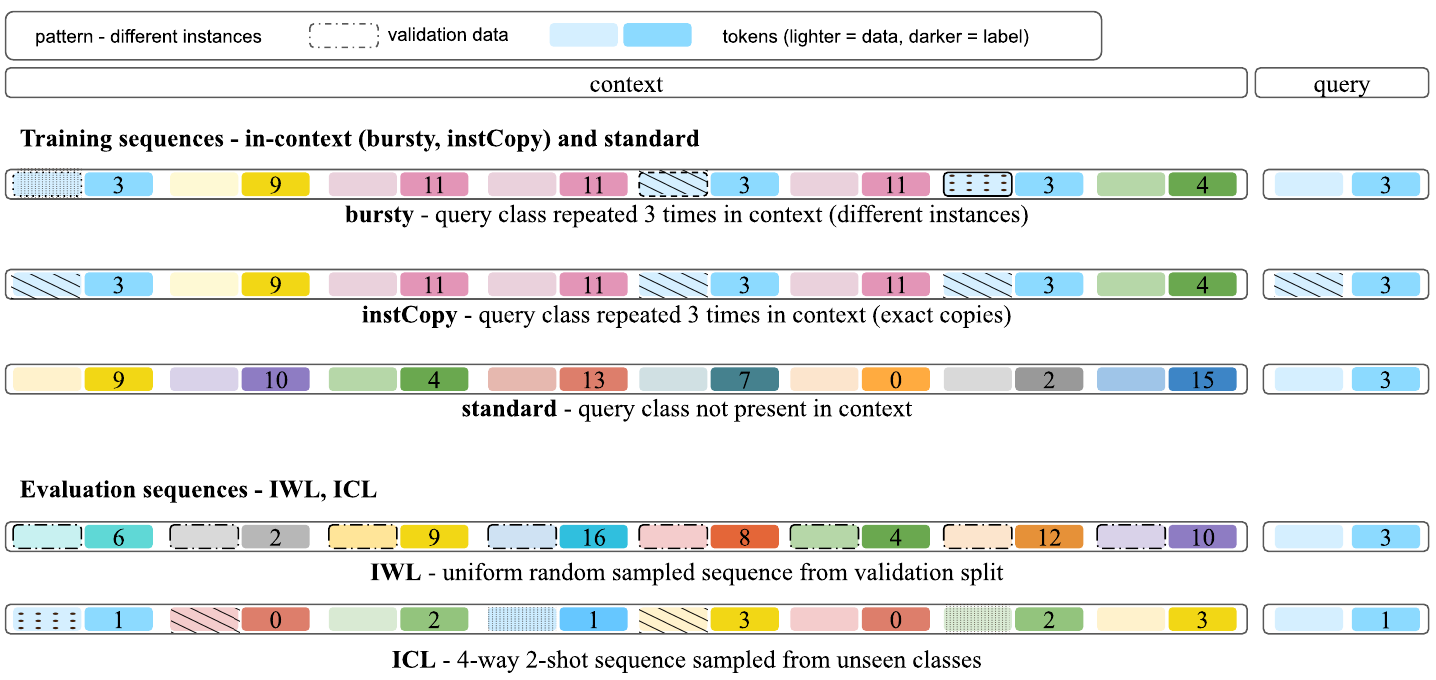}
    \caption{Different training and evaluation sequences with the main difference being the number of repetitions and the use of identical copies in the context. 
    }
    \label{fig:sequences_overview}
\end{figure}

We illustrate the difference between the sequences leveraged in our experimental setup on \cref{fig:sequences_overview}. We employ a sequence of length $L=8$ with eight image-label pairs. In this case, standard sequences have 8 unique image-label pairs in the context, and the 9th image comes from a 9th class. For bursty sequences, we distinguish between high burstiness in the sequence (referred to as \textbf{bursty sequence}) with three instances from the query class in the sequence and low burstiness (referred to as \textbf{bursty (low) sequence}) with one example from the query class in the sequence. 
We further introduce bursty sequence \textbf{instCopy}, which follows the same logic as bursty sequence, but instead of having three instances from the query class in the sequence, it has the same example as the query image repeated (copy-pasted) three times in the sequence (see the same pattern in query-class instances on \cref{fig:sequences_overview}). We introduce this type of sequence motivated by the frequent repetitions in natural language.  

\textbf{Evaluation sequences.} During the evaluation, we leverage 2 different types -- IWL and ICL evaluation sequences. 
IWL is evaluated for the multi-class classification task on the held-out samples from the training classes. The standard sequences, with an uniformly sampled format, are used for IWL evaluation (see Figure~\ref{fig:sequences_overview}). 
ICL is evaluated in a few-shot classification setting for 2-way-4-shot and 4-way-2-shot tasks. We present results in the main paper for the more challenging 4-way-2-shot setting, while 2-way 4-shot results are included in \cref{app:datasets_details}. This evaluation is performed on held-out novel classes. The trained classifier output is used for the few-shot evaluation, utilising label mappings from 0-1 or 0-3, to 2-way 4-shot and 4-way 2-shot evaluation, respectively.

\textbf{Dataset construction.}
We conduct our controlled experiments and analysis on the Omniglot dataset~\cite{omniglot} and scale to the more realistic visual dataset, often used in few-shot learning evaluation: CIFAR-100 ~\cite{cifarfs}, Caltech-101 ~\cite{caltech}, and DTD texture datasets ~\cite{dtd}. Omniglot contains 1623 classes with 20 images each, following previous work~\cite{NEURIPS2022_77c6ccac}, we use 1600 classes for training and the remaining 23 as novel classes for ICL evaluation. For CIFAR-100, Caltech-101 and DTD datasets we perform ICL evaluation using 20, 10, and 10 novel classes, respectively.
More experimental details are included in the~\cref{app:experiment_details}.

\section{How to enable ICL? }\label{sec:how_to_enable_icl} 

Prior work has identified specific circuits in transformer models as the working mechanism of ICL -- induction heads~\cite{olsson2022context, Reddy2024the, singh2024needs}. The induction head embodies the core concept of in-context learning: examining the context to identify the most similar or relevant token and then retrieving the associated, already aggregated knowledge. ICL require two underlying components to be established: a similarity or look-up function and a head attending to the previous token. These two components are mutually dependent -- the similarity function is ineffective without meaningful aggregated information by the previous-token head, and the previous-token cannot be optimized without a similarity mechanism to retrieve and apply the stored information. 

\subsection{Why is ICL learned and non-transient on text but not on visual data?}

The presence of ICL capabilities in LLMs is well established. However, prior work has struggled to obtain stable ICL in other domains, such as vision.  Chen et al.~\cite{NEURIPS2022_77c6ccac} demonstrated that burstiness in the training sequences and skewness in data -- both inherently present in natural language -- show the emergence of ICL in simple visual tasks using the Omniglot dataset. However, despite these changes, the vision model still suffers from diminished IWL performance along with a transient decline in ICL performance as training progresses.

\begin{figure}[tp!]
    \centering
        \includegraphics[width=0.9\textwidth]{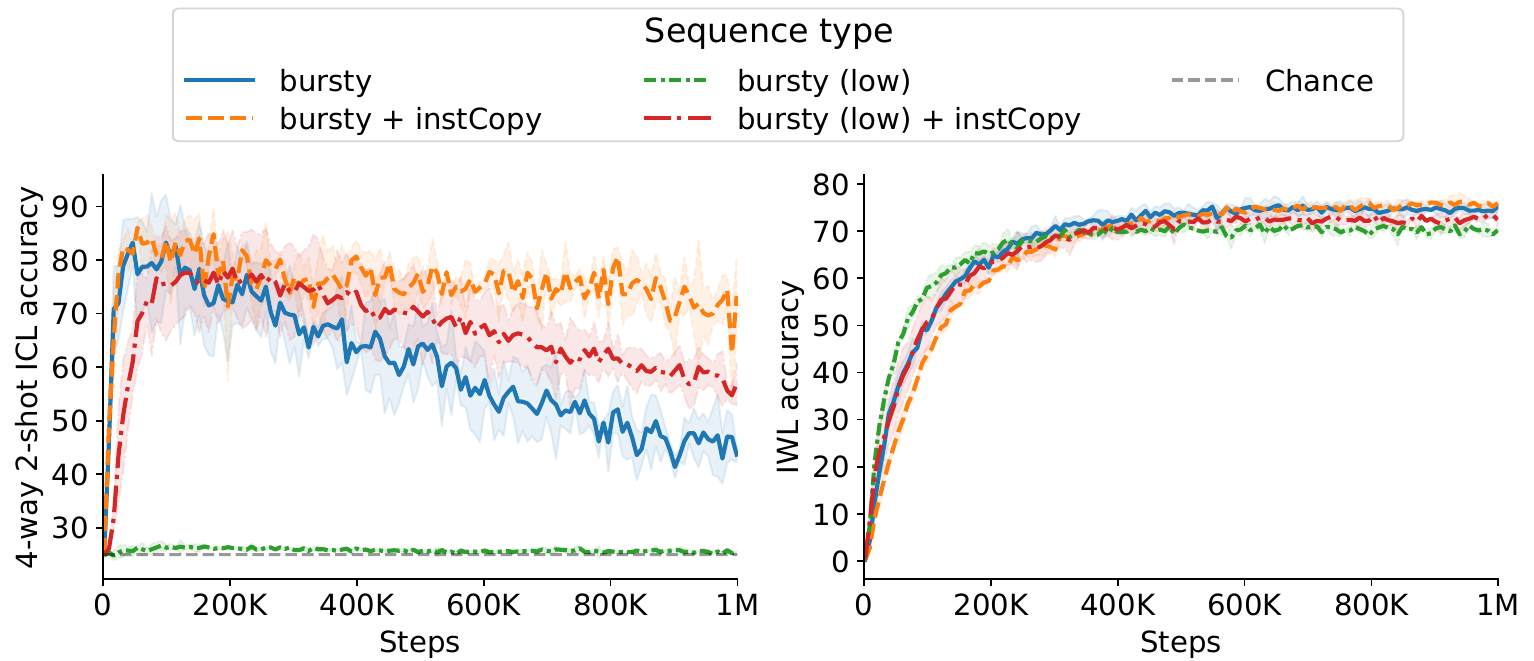}
        \caption{Exact copies in the context (instCopy) promote ICL performance and reduce transiency. Only a single copy ensures ICL emergence (bursty (low) case).}
    \label{fig:repetition_skewness}
\end{figure}

Besides burstiness and skewness, we argue that natural language typically contains many exact token copies and n-grams (as shown in~\cref{app:burstiness}), which, we believe, is an important factor for stable and non-transient ICL in LLMs. We train the model conforming to the bursty sequences introduced by Chan et al.~\cite{NEURIPS2022_77c6ccac} and propose a new type of bursty sequence with exact instance copies in the context (instCopy) to test this hypothesis.

From~\cref{fig:repetition_skewness}, we observe that including bursty sequences in the training data indeed leads to the emergence of the ICL, which supports previous works~\cite{NEURIPS2022_77c6ccac, singh2023the, singh2024needs}. However, the model only achieves strong and more stable (less transient) ICL performance while using exact copies in the bursty sequences (instCopy). Furthermore, we can see that high burstiness is not essential for ICL -- a single exact copy in the context (bursty (low)+instCopy) is sufficient to obtain ICL. \textbf{This confirms that exact copies are a stronger driving factor for ICL, even surpassing the burstiness, as previously reported }~\cite{NEURIPS2022_77c6ccac}.

\subsection{Why do exact copies help?}\label{sec:exact_copy_help}

In~\cref{fig:repetition_skewness}, we show that exact copies facilitate strong and stable ICL performance. We argue that this is due to the simplified similarity function, which breaks its interdependence with previous-token head learning and ensures that the model prioritizes the formation of the previous-token head. 

To confirm this argument, we compare the QK attention scores of the model trained with bursty sequences and the model trained with combined burstiness and exact copies (instCopy) as in-context sequences.

\begin{figure}[!tp]
    \centering
    \includegraphics[width=\textwidth]{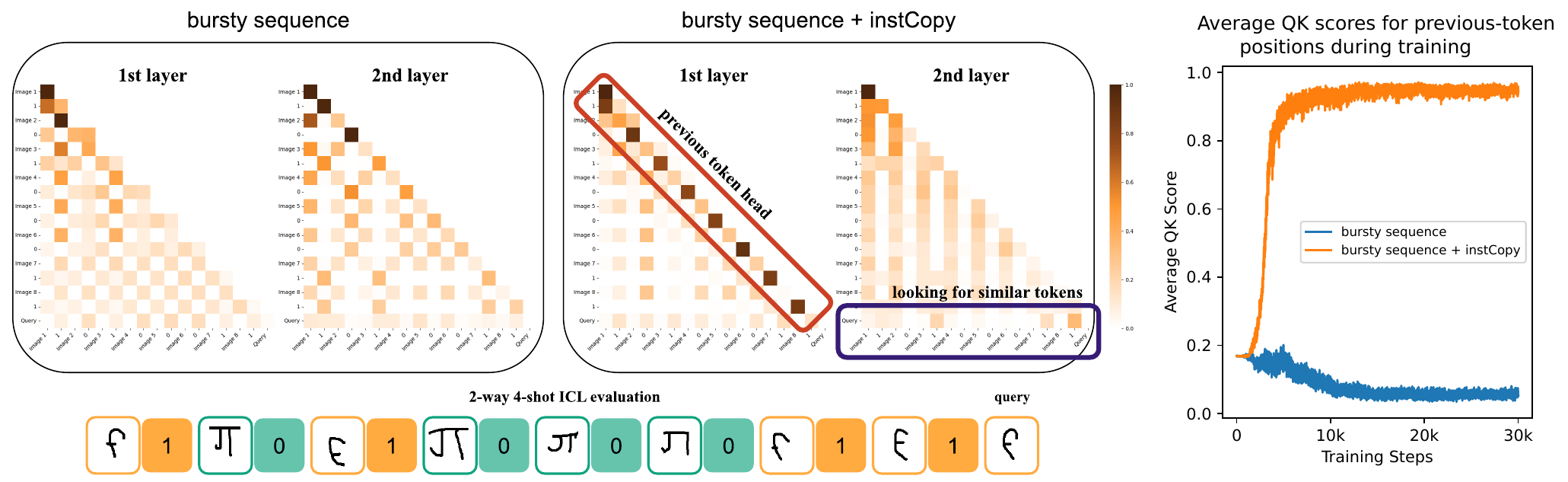}
    \caption{We observe clear induction head and ICL emergence during inference only for the model trained with burstiness and exact copies (bursty + InstCopy). Attention patterns in the QK space reveal a previous-token head in layer one (diagonal with offset 1) and a query token attending to the most similar label tokens in layer two. On the right, we show average QK scores over the previous-token head positions for these models during training. High attention scores for instCopy sequences confirms previous-token head formation.
    }
    \label{fig:ih_analysis}
\end{figure}
 
During training, we observe higher scores for tokens corresponding to the query label and formation of previous-token heads only for the model with instCopy sequences. In~\cref{fig:ih_analysis}(right), we trace the formation of previous-token heads during training by computing the averaged QK values off the diagonal (expected positions for previous-token head) over the training process. The previous token heads indicate aggregation of knowledge from the image to the label token. Since the similarity function is now trivial, the model learns to attend query to previous tokens and successfully perform the needed ICL operations.

We observe the same patterns during inference for 2-way 4-shot classification on novel classes, as illustrated in~\cref{fig:ih_analysis} (left). We observe ICL performance only for the model trained with bursty sequences and exact copies (bursty + instCopy). For the same model, we observe more attention between similar tokens in the sequence and a visible previous-token head. This confirms that \textbf{including exact copies in the context indeed simplifies the learning of the similarity function and promotes the formation of a previous-token head, which is then utilised during inference to make a correct ICL prediction.}

\subsection{What unlocks ICL for various visual datasets?}\label{sec:analysis_other_data}

\begin{figure}[bp!]
    \centering
        \includegraphics[width=0.8\textwidth]{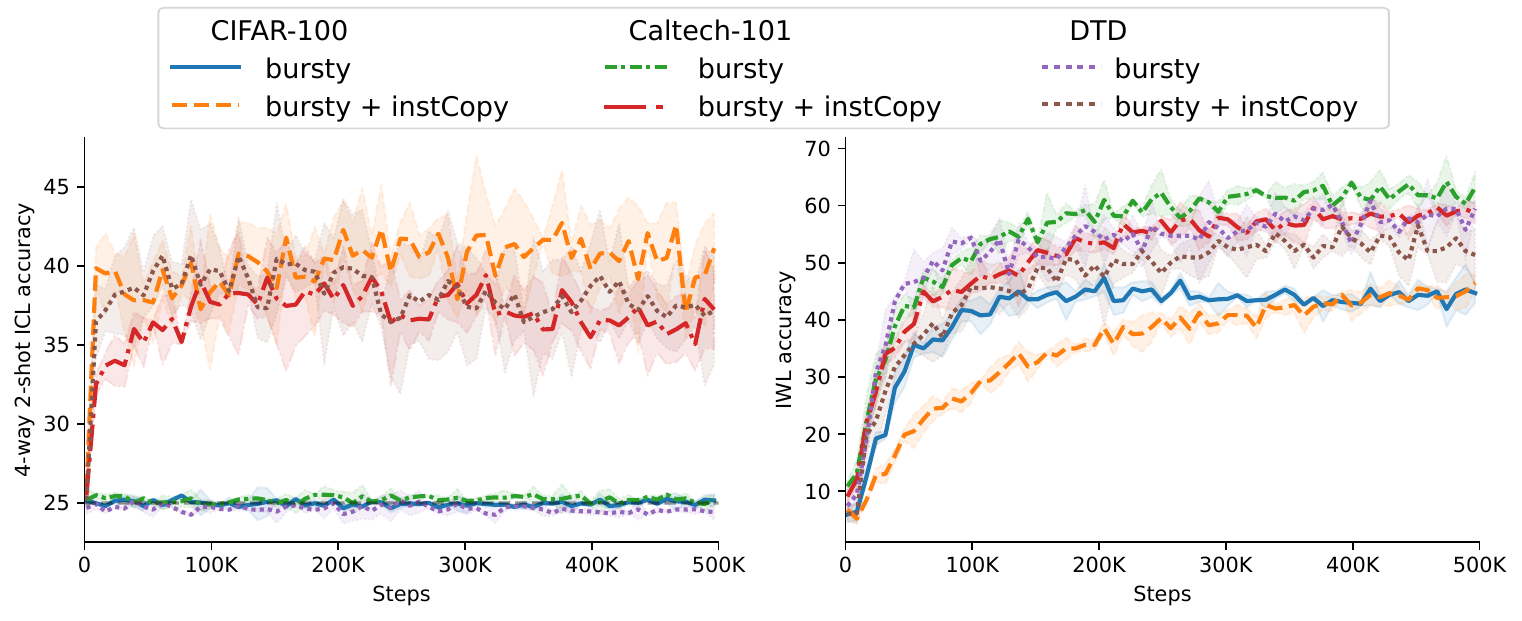}
    \caption{Only when employing exact copies in the context (instCopy), we ensure ICL emergence on the image classification datasets Cifar, Caltech and DTD.}
    \label{fig:other_datasets}
\end{figure}
Previously, we confirmed that bursty sequences (without exact copies) unlock in-context learning on simple vision datasets like Omniglot. However, we find that the same setup fails to obtain any ICL for more complex vision datasets like DTD~\cite{dtd}, CIFAR-100~\cite{cifarfs}, and Caltech-101~\cite{caltech}, as shown in~\ref{fig:other_datasets}.

We hypothesize that burstiness alone does not provide a sufficient signal to learn the similarity function, which is also much harder for complex images. Here, we include exact instance copies (instCopy) in the bursty sequences.
We observe that \textbf{instCopy enables strong ICL performance for all three complex visual datasets}, as shown in~\cref{fig:other_datasets}. We also observe that the IWL performance remains largely unaffected.

\subsection{Does in-weight learning (IWL) task influence ICL?}\label{sec:iwl_other}

Language modelling is a significantly harder task than Omniglot classification, yet we naturally observe strong ICL performance in LLMs, but not in Omniglot. Does the IWL task influence ICL? If yes, then how?

The emergence of ICL requires the formation of induction heads. However, when the IWL task is overly simple, the model can exhibit simplicity bias and prioritizes IWL learning over learning of induction heads. Even highly bursty in-context sequences may fail to enable ICL in such cases. Following this, we argue that an in-weight task must have a minimum level of complexity to encourage ICL.

\begin{figure}[bp!]
    \centering
    \includegraphics[width=\linewidth]{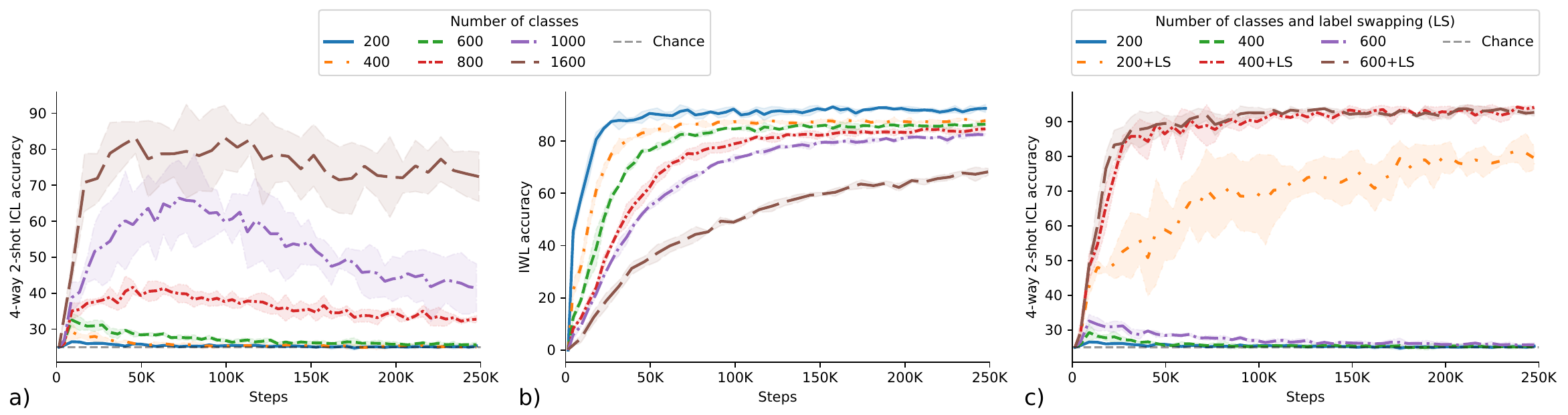}
    \caption{When increasing the number of classes monotonically (a, b) or applying label swapping (c), ICL performance improves as the IWL objective gets harder. }
    \label{fig:iwl_classes_noise}
\end{figure}

To test this hypothesis, we make IWL tasks more challenging in two ways -- increasing the number of classes and adding label noise by label swapping. In~\cref{app:iwl_task} we demonstrate two more ways.

\textbf{Number of training classes vs. ICL.} 
We create IWL tasks on Omniglot dataset with an increasing number of classes varying from 200 to 1600 classes. Using the training setup with bursty in-context sequences (without instCopy), we evaluate ICL and IWL performance. 
In~\cref {fig:iwl_classes_noise}ab, we observe that IWL converges more slowly as the number of classes increases, indicating that IWL task becomes more challenging with a larger class set. In contrast, ICL performance improves when the number of classes increase. This trend suggests a competition between IWL and ICL circuits in the early phase of training. If the IWL task is too simple, it may fulfil the IWL objective without learning the ICL mechanism. Prior work~\cite{NEURIPS2022_77c6ccac,Reddy2024the} indicate similar findings. However, they attribute this improvement to the presence of many rarely occurring classes, which can be interpreted as another way to make the IWL task harder.
This shows that \textbf{increasing the number of classes makes IWL harder and improves ICL}.
 
\textbf{Label noise vs. ICL.} 
Here, we create different training setups with label noise using Omniglot~\cite{omniglot}.
We perform random label swapping, where labels of the query items are randomly assigned to another training class in 20\% of all sequences. For the case of a bursty sequence, we change the label mappings to all repetitions in sequences belonging to the query class. We train models with 200, 400, and 600 classes, where no ICL was observed without noise, shown in~\cref{fig:iwl_classes_noise}ab, due to the overly simple IWL.
In~\cref{fig:iwl_classes_noise}c, we observe that label swapping significantly improves ICL performance while for IWL, we observe similar trend of slower convergence as in~\cref{fig:iwl_classes_noise}b (additional results in ~\cref{app:iwl_task}). 
Label swapping makes the IWL task challenging due to the introduced label noise in the standard sequences. In contrast, label swapping promotes in-context mechanisms in bursty sequences since the model cannot rely on in-weight class embeddings to minimize the loss.
We observe strong ICL performance even for the simple IWL case of 200 classes with added label swapping. This further confirms that \textbf{hard IWL task with label noise via label swapping leads to ICL}.

\section{Enabling ICL for EEG classification} \label{sec:eeg}

As mentioned earlier, ICL ensures fast adaptation to new tasks, algorithms, and unseen scenarios -- a desirable feature for many applications. 
Enabling ICL can be challenging for real-world datasets, which are often noisy or exhibit significant variance between instances -- a common problem in EEG data. 
On the other hand, EEG tasks would greatly benefit from ICL ability, as it would enable fast adaptation to novel datasets and setups without the need for retraining, which is currently not the case~\cite{patil2024coordconformer, duan2020zero}. Following this motivation, we aim to enable ICL for EEG data, a modality more challenging and noisy than text or images.

We modify our initial setup from image classification to EEG classification (experimental details in~\cref{app:eeg_experimental}) and attempt to enable ICL by relying solely on burstiness following~\cite{NEURIPS2022_77c6ccac}. However, similar to image classification results in~\cref{sec:analysis_other_data} -- we observe no ICL emergence. This suggests that enabling ICL for EEG classification requires further interventions to help the model overcome the learning of the similarity function and form the previous-token head: exact insights we have provided in~\cref{sec:how_to_enable_icl}.

\textbf{Enabling ICL for EEG.}
We build on our insights from~\cref{sec:exact_copy_help} and~\cref{sec:analysis_other_data} about the importance of the similarity function and previous-token head for ICL. 
Given the high noise and variability in EEG data, we posit that using exact copies can help the model to initially bypass the complex similarity function and more effectively learn the previous-token head.  We further investigate the relationship between the difficulty of IWL task and ICL emergence. Our EEG setup doesn't allow us many base classes for IWL task, therefore, we only employ label swapping to make IWL task harder to promote ICL. Details about the training setup and sequence construction are provided in~\cref{app:eeg_train_seq}.

\textbf{Results.} 
In~\cref{tab:eeg_results}, we present the results for ICL performance on three novel datasets: BNCI~\cite{bnci2012}, HGD~\cite{schirrmeister2017}, and Zhou~\cite{zhou2016} while using a mixture of datasets for training (details in~\cref{app:eeg_experimental}). We compare three models: 1) a baseline bursty model trained with a combination of bursty in-context and standard sequences only, 2) a model with burstiness and label swapping method, and finally, 3) a model employing burstiness with exact copies and label swapping. 

Consistent with the image classification results, we observe no ICL performance when using only a combination of bursty in-context and standard sequences. Utilizing burstiness with label swapping yields little to no ICL. However, \textbf{ICL reliably emerges for EEG data when both bursty sequences with exact token copies and label swapping are applied}, as this provides a balance that supports both ICL and IWL learning. This further supports our insights into the training dynamics for ICL emergence; they not only apply to diverse image datasets but also in more real-world domains such as EEG.

\begin{table}[tp]
\centering
\begin{tabular}{@{\hspace{0.15cm}} c @{\hspace{0.7cm}}c @{\hspace{0.7cm}}c @{\hspace{0.25cm}}|@{\hspace{0.25cm}}c @{\hspace{0.35cm}}c @{\hspace{0.35cm}}c @{\hspace{0.15cm}}}
\hline
burstiness & label swapping & instCopy & BNCI~\cite{bnci2012} & HGD~\cite{schirrmeister2017} & Zhou~\cite{zhou2016} \\ \hline
\checkmark&  &  & 33.49 & 33.24 & 33.82 \\
\checkmark & \checkmark &  & 35.15 & 34.75 & 38.82 \\
\checkmark  & \checkmark  & \checkmark  & \textbf{37.56} & \textbf{39.65} & \textbf{47.66} \\ \hline
\end{tabular}
\caption{ICL generalization results across different novel datasets with random chance of 33\%. We observe the best ICL emergence when exact copies are present in the context (instCopy), and label swapping has made the IWL task harder.
}
\label{tab:eeg_results}
\end{table}

\section{Discussion}\label{sec:discussion}

\textbf{Key insights.} In this work, we demonstrate how to unlock ICL for various modalities beyond text, providing novel insights into the training dynamics of ICL. Specifically, we demonstrate that ICL can be learned more easily by breaking the interdependence between the two operations necessary for ICL: a similarity function that matches the relevant tokens with the query and a previous-token head for knowledge aggregation. 

We confirm earlier works~\cite{NEURIPS2022_77c6ccac, Reddy2024the, singh2023the} on the importance of data distributional properties coming from natural language for ICL emergence. However, we find that using exact token copies during training facilitates stronger in-context learning, leading to higher accuracy and more stable results. We further show that, against prior beliefs~\cite {NEURIPS2022_77c6ccac}, burstiness is not essential for ICL -- a single token copy in the context can be sufficient for ICL emergence. We provide an explanation and evidence of why exact token copies facilitate ICL emergence: they simplify the similarity function to be learned, breaking the interdependence of ICL learning mechanisms and allowing the formation of previous-token heads.

We further identify another strong driver for ICL emergence -- the relationship between ICL and IWL task difficulty. When the IWL task is more challenging, the model is more likely to rely on context and learn ICL.
Finally, we confirm our novel insights
by demonstrating that exact token copies and increased task difficulty unlock ICL performance across various visual datasets, where previous findings failed to do so~\cite{NEURIPS2022_77c6ccac}. We even enable ICL on much more complex and noisy continuous data, such as EEG, where ICL now, for the first time, allows few-shot transfer to novel datasets.

\textbf{Limitations and Future work.} ICL performance exhibits high variance when trained with simple IWL tasks, probably due to its sensitivity to training sequences~\cite{press-etal-2023-measuring}. Furthermore, we observe a significant impact on ICL stability due to certain model design choices. Promising future research directions include improving robustness and expanding our training insights to additional applications beyond image and EEG classification.

\section{Acknowledgments}
This research was funded by the German Research Foundation (DFG) 417962828, 539134284 and 499552394 (SFB 1597 - Small Data), and by the German Ministry for Economy and Climate Protection via a decision by the German parliament (19A23014R).

%
%
%
%
\bibliographystyle{splncs04}
\bibliography{egbib}

\begin{thebibliography}{10}
\providecommand{\url}[1]{\texttt{#1}}
\providecommand{\urlprefix}{URL }
\providecommand{\doi}[1]{https://doi.org/#1}

\bibitem{agarwal2024manyshot}
Agarwal, R., Singh, A., Zhang, L.M., Bohnet, B., Rosias, L., Chan, S.C., Zhang, B., Anand, A., Abbas, Z., Nova, A., Co-Reyes, J.D., Chu, E., Behbahani, F., Faust, A., Larochelle, H.: Many-shot in-context learning. In: The Thirty-eighth Annual Conference on Neural Information Processing Systems (2024), \url{https://openreview.net/forum?id=AB6XpMzvqH}

\bibitem{akyurek2023what}
Aky{\"u}rek, E., Schuurmans, D., Andreas, J., Ma, T., Zhou, D.: What learning algorithm is in-context learning? investigations with linear models. In: The Eleventh International Conference on Learning Representations (2023), \url{https://openreview.net/forum?id=0g0X4H8yN4I}

\bibitem{akyurek2024incontext}
Akyürek, E., Wang, B., Kim, Y., Andreas, J.: In-context language learning: Architectures and algorithms (2024)

\bibitem{cutcat}
Al-Saegh, A., Dawwd, S., Abdul-Jabbar, J.: Cutcat: An augmentation method for eeg classification. Neural Networks  \textbf{141} (06 2021). \doi{10.1016/j.neunet.2021.05.032}

\bibitem{alayrac2022flamingo}
Alayrac, J.B., Donahue, J., Luc, P., Miech, A., Barr, I., Hasson, Y., Lenc, K., Mensch, A., Millican, K., Reynolds, M., Ring, R., Rutherford, E., Cabi, S., Han, T., Gong, Z., Samangooei, S., Monteiro, M., Menick, J., Borgeaud, S., Brock, A., Nematzadeh, A., Sharifzadeh, S., Binkowski, M., Barreira, R., Vinyals, O., Zisserman, A., Simonyan, K.: Flamingo: a visual language model for few-shot learning. In: Advances in Neural Information Processing Systems (2022), \url{https://openreview.net/forum?id=EbMuimAbPbs}

\bibitem{NEURIPS2023_b2e63e36}
Bai, Y., Chen, F., Wang, H., Xiong, C., Mei, S.: Transformers as statisticians: Provable in-context learning with in-context algorithm selection. In: Advances in Neural Information Processing Systems. vol.~36, pp. 57125--57211 (2023), \url{https://proceedings.neurips.cc/paper_files/paper/2023/file/b2e63e36c57e153b9015fece2352a9f9-Paper-Conference.pdf}

\bibitem{bai2023sequentialmodelingenablesscalable}
Bai, Y., Geng, X., Mangalam, K., Bar, A., Yuille, A.L., Darrell, T., Malik, J., Efros, A.A.: Sequential modeling enables scalable learning for large vision models. In: 2024 IEEE/CVF Conference on Computer Vision and Pattern Recognition (CVPR). pp. 22861--22872 (2024). \doi{10.1109/CVPR52733.2024.02157}

\bibitem{visprompt}
Bar, A., Gandelsman, Y., Darrell, T., Globerson, A., Efros, A.A.: Visual prompting via image inpainting. In: Oh, A.H., Agarwal, A., Belgrave, D., Cho, K. (eds.) Advances in Neural Information Processing Systems (2022), \url{https://openreview.net/forum?id=o4uFFg9_TpV}

\bibitem{cifarfs}
Bertinetto, L., Henriques, J.F., Torr, P., Vedaldi, A.: Meta-learning with differentiable closed-form solvers. In: International Conference on Learning Representations (2019), \url{https://openreview.net/forum?id=HyxnZh0ct7}

\bibitem{NEURIPS2020_1457c0d6}
Brown, T., Mann, B., Ryder, N., Subbiah, M., Kaplan, J.D., Dhariwal, P., Neelakantan, A., Shyam, P., Sastry, G., Askell, A., Agarwal, S., Herbert-Voss, A., Krueger, G., Henighan, T., Child, R., Ramesh, A., Ziegler, D., Wu, J., Winter, C., Hesse, C., Chen, M., Sigler, E., Litwin, M., Gray, S., Chess, B., Clark, J., Berner, C., McCandlish, S., Radford, A., Sutskever, I., Amodei, D.: Language models are few-shot learners. In: Advances in Neural Information Processing Systems. vol.~33, pp. 1877--1901 (2020), \url{https://proceedings.neurips.cc/paper_files/paper/2020/file/1457c0d6bfcb4967418bfb8ac142f64a-Paper.pdf}

\bibitem{NEURIPS2022_77c6ccac}
Chan, S., Santoro, A., Lampinen, A., Wang, J., Singh, A., Richemond, P., McClelland, J., Hill, F.: Data distributional properties drive emergent in-context learning in transformers. In: Advances in Neural Information Processing Systems. vol.~35, pp. 18878--18891 (2022)

\bibitem{chen2024parallel}
Chen, Y., Zhao, C., Yu, Z., McKeown, K., He, H.: Parallel structures in pre-training data yield in-context learning (2024), \url{https://arxiv.org/abs/2402.12530}

\bibitem{dtd}
Cimpoi, M., Maji, S., Kokkinos, I., Mohamed, S., , Vedaldi, A.: Describing textures in the wild. In: Proceedings of the {IEEE} Conf. on Computer Vision and Pattern Recognition ({CVPR}) (2014)

\bibitem{clark-etal-2019-bert}
Clark, K., Khandelwal, U., Levy, O., Manning, C.D.: What does {BERT} look at? an analysis of {BERT}{'}s attention. In: Linzen, T., Chrupa{\l}a, G., Belinkov, Y., Hupkes, D. (eds.) Proceedings of the 2019 ACL Workshop BlackboxNLP: Analyzing and Interpreting Neural Networks for NLP. pp. 276--286. Association for Computational Linguistics, Florence, Italy (Aug 2019). \doi{10.18653/v1/W19-4828}, \url{https://aclanthology.org/W19-4828/}

\bibitem{dai-etal-2023-gpt}
Dai, D., Sun, Y., Dong, L., Hao, Y., Ma, S., Sui, Z., Wei, F.: Why can {GPT} learn in-context? language models secretly perform gradient descent as meta-optimizers. In: Findings of the Association for Computational Linguistics: ACL 2023. pp. 4005--4019 (2023)

\bibitem{duan2020zero}
Duan, L., Li, J., Ji, H., Pang, Z., Zheng, X., Lu, R., Li, M., Zhuang, J.: Zero-shot learning for eeg classification in motor imagery-based bci system. IEEE Transactions on Neural Systems and Rehabilitation Engineering  \textbf{28}(11),  2411--2419 (2020)

\bibitem{NEURIPS2024_icl_markov}
Edelman, E., Tsilivis, N., Edelman, B.L., Malach, E., Goel, S.: The evolution of statistical induction heads: In-context learning markov chains. In: Globerson, A., Mackey, L., Belgrave, D., Fan, A., Paquet, U., Tomczak, J., Zhang, C. (eds.) Advances in Neural Information Processing Systems. vol.~37, pp. 64273--64311. Curran Associates, Inc. (2024), \url{https://proceedings.neurips.cc/paper_files/paper/2024/file/75b0edb869e2cd509d64d0e8ff446bc1-Paper-Conference.pdf}

\bibitem{elhelo2025inferringfunctionalityattentionheads}
Elhelo, A., Geva, M.: Inferring functionality of attention heads from their parameters (2025), \url{https://arxiv.org/abs/2412.11965}

\bibitem{caltech}
Fei-Fei, L., Fergus, R., Perona, P.: Learning generative visual models from few training examples: An incremental bayesian approach tested on 101 object categories. In: 2004 Conference on Computer Vision and Pattern Recognition Workshop. pp. 178--178 (2004)

\bibitem{wikidump}
Foundation, W.: Wikimedia downloads, \url{https://dumps.wikimedia.org}

\bibitem{NEURIPS2022_c529dba0}
Garg, S., Tsipras, D., Liang, P.S., Valiant, G.: What can transformers learn in-context? a case study of simple function classes. In: Advances in Neural Information Processing Systems. vol.~35, pp. 30583--30598 (2022)

\bibitem{Gokaslan2019OpenWeb}
Gokaslan, A., Cohen, V., Pavlick, E., Tellex, S.: Openwebtext corpus. \url{http://Skylion007.github.io/OpenWebTextCorpus} (2019)

\bibitem{physionet2000}
Goldberger, A.L., Amaral, L.A.N., Glass, L., Hausdorff, J.M., Ivanov, P.C., Mark, R.G., Mietus, J.E., Moody, G.B., Peng, C.K., Stanley, H.E.: Physiobank, physiotoolkit, and physionet. Circulation  \textbf{101}(23),  e215--e220 (2000). \doi{10.1161/01.CIR.101.23.e215}, \url{https://www.ahajournals.org/doi/abs/10.1161/01.CIR.101.23.e215}

\bibitem{gu2023pretraininglearncontext}
Gu, Y., Dong, L., Wei, F., Huang, M.: Pre-training to learn in context (2023), \url{https://arxiv.org/abs/2305.09137}

\bibitem{han-etal-2023-understanding}
Han, X., Simig, D., Mihaylov, T., Tsvetkov, Y., Celikyilmaz, A., Wang, T.: Understanding in-context learning via supportive pretraining data. In: Rogers, A., Boyd-Graber, J., Okazaki, N. (eds.) Proceedings of the 61st Annual Meeting of the Association for Computational Linguistics (Volume 1: Long Papers) (Jul 2023)

\bibitem{he2016deep}
He, K., Zhang, X., Ren, S., Sun, J.: Deep residual learning for image recognition. In: Proceedings of the IEEE conference on computer vision and pattern recognition. pp. 770--778 (2016)

\bibitem{hollmann2023tabpfn}
Hollmann, N., M{\"u}ller, S., Eggensperger, K., Hutter, F.: Tabpfn: A transformer that solves small tabular classification problems in a second. In: International Conference on Learning Representations 2023 (2023)

\bibitem{hollmann2025tabpfn}
Hollmann, N., M{\"u}ller, S., Purucker, L., Krishnakumar, A., K{\"o}rfer, M., Hoo, S.B., Schirrmeister, R.T., Hutter, F.: Accurate predictions on small data with a tabular foundation model. Nature  (01 2025). \doi{10.1038/s41586-024-08328-6}, \url{https://www.nature.com/articles/s41586-024-08328-6}

\bibitem{huang2024multimodaltaskvectorsenable}
Huang, B., Mitra, C., Arbelle, A., Karlinsky, L., Darrell, T., Herzig, R.: Multimodal task vectors enable many-shot multimodal in-context learning (2024), \url{https://arxiv.org/abs/2406.15334}

\bibitem{jiang2024large}
Jiang, W.B., Zhao, L.M., Lu, B.L.: Large brain model for learning generic representations with tremendous eeg data in bci. arXiv preprint arXiv:2405.18765  (2024)

\bibitem{Kingma2014AdamAM}
Kingma, D.P., Ba, J.: Adam: A method for stochastic optimization. arXiv preprint arXiv:1412.6980  (2014)

\bibitem{omniglot}
Lake, B.M., Salakhutdinov, R., Tenenbaum, J.B.: Human-level concept learning through probabilistic program induction. Science  \textbf{350}(6266),  1332--1338 (2015)

\bibitem{laur2023idefics}
Laurençon, H., Saulnier, L., Tronchon, L., Bekman, S., Singh, A., Lozhkov, A., Wang, T., Karamcheti, S., Rush, A.M., Kiela, D., Cord, M., Sanh, V.: Obelics: An open web-scale filtered dataset of interleaved image-text documents (2023), \url{https://arxiv.org/abs/2306.16527}

\bibitem{levine2022pretrainig}
Levine, Y., Wies, N., Jannai, D., Navon, D., Hoshen, Y., Shashua, A.: The inductive bias of in-context learning: Rethinking pretraining example design. In: International Conference on Learning Representations (2022), \url{https://openreview.net/forum?id=lnEaqbTJIRz}

\bibitem{li5177120two}
Li, L., Wei, B.: A two-stage eeg zero-shot classification algorithm guided by class reconstruction. Available at SSRN 5177120  (2025)

\bibitem{lindsey2025biology}
Lindsey, J., Gurnee, W., Ameisen, E., Chen, B., Pearce, A., Turner, N.L., Citro, C., Abrahams, D., Carter, S., Hosmer, B., Marcus, J., Sklar, M., Templeton, A., Bricken, T., McDougall, C., Cunningham, H., Henighan, T., Jermyn, A., Jones, A., Persic, A., Qi, Z., Thompson, T.B., Zimmerman, S., Rivoire, K., Conerly, T., Olah, C., Batson, J.: On the biology of a large language model. Transformer Circuits Thread  (2025), \url{https://transformer-circuits.pub/2025/attribution-graphs/biology.html}

\bibitem{liu-etal-2022-makes}
Liu, J., Shen, D., Zhang, Y., Dolan, B., Carin, L., Chen, W.: What makes good in-context examples for {GPT}-3? In: Proceedings of Deep Learning Inside Out (DeeLIO 2022): The 3rd Workshop on Knowledge Extraction and Integration for Deep Learning Architectures (May 2022)

\bibitem{liu2024incontextvectorsmakingcontext}
Liu, S., Ye, H., Xing, L., Zou, J.: In-context vectors: Making in context learning more effective and controllable through latent space steering (2024), \url{https://arxiv.org/abs/2311.06668}

\bibitem{liu2302brainclip}
Liu, Y., Ma, Y., Zhou, W., Zhu, G., Zheng, N.: Brainclip: Bridging brain and visual-linguistic representation via clip for generic natural visual stimulus decoding. arxiv 2023. arXiv preprint arXiv:2302.12971  (2023)

\bibitem{long-etal-2023-adapt}
Long, Q., Wang, W., Pan, S.: Adapt in contexts: Retrieval-augmented domain adaptation via in-context learning. In: Bouamor, H., Pino, J., Bali, K. (eds.) Proceedings of the 2023 Conference on Empirical Methods in Natural Language Processing. pp. 6525--6542. Association for Computational Linguistics, Singapore (Dec 2023). \doi{10.18653/v1/2023.emnlp-main.402}, \url{https://aclanthology.org/2023.emnlp-main.402/}

\bibitem{Loshchilov2017DecoupledWD}
Loshchilov, I., Hutter, F.: Decoupled weight decay regularization. In: International Conference on Learning Representations (2017), \url{https://api.semanticscholar.org/CorpusID:53592270}

\bibitem{wordvec_neurips}
Mikolov, T., Sutskever, I., Chen, K., Corrado, G.S., Dean, J.: Distributed representations of words and phrases and their compositionality. In: Burges, C., Bottou, L., Welling, M., Ghahramani, Z., Weinberger, K. (eds.) Advances in Neural Information Processing Systems. vol.~26. Curran Associates, Inc. (2013), \url{https://proceedings.neurips.cc/paper_files/paper/2013/file/9aa42b31882ec039965f3c4923ce901b-Paper.pdf}

\bibitem{min-etal-2022-metaicl}
Min, S., Lewis, M., Zettlemoyer, L., Hajishirzi, H.: {M}eta{ICL}: Learning to learn in context. In: Carpuat, M., de~Marneffe, M.C., Meza~Ruiz, I.V. (eds.) Proceedings of the 2022 Conference of the North American Chapter of the Association for Computational Linguistics: Human Language Technologies. pp. 2791--2809. Association for Computational Linguistics, Seattle, United States (Jul 2022). \doi{10.18653/v1/2022.naacl-main.201}, \url{https://aclanthology.org/2022.naacl-main.201/}

\bibitem{olsson2022context}
Olsson, C., Elhage, N., Nanda, N., Joseph, N., DasSarma, N., Henighan, T., Mann, B., Askell, A., Bai, Y., Chen, A., Conerly, T., Drain, D., Ganguli, D., Hatfield-Dodds, Z., Hernandez, D., Johnston, S., Jones, A., Kernion, J., Lovitt, L., Ndousse, K., Amodei, D., Brown, T., Clark, J., Kaplan, J., McCandlish, S., Olah, C.: In-context learning and induction heads. Transformer Circuits Thread  (2022), https://transformer-circuits.pub/2022/in-context-learning-and-induction-heads/index.html

\bibitem{patil2024coordconformer}
Patil, S., Schirrmeister, R.T., Hutter, F., Ball, T.: Coordconformer: Heterogenous {EEG} datasets decoding using transformers. In: ICML 2024 Workshop on Geometry-grounded Representation Learning and Generative Modeling (2024), \url{https://openreview.net/forum?id=RlYWwZXlsJ}

\bibitem{pawelczyk2024icl_unlearning}
Pawelczyk, M., Neel, S., Lakkaraju, H.: In-context unlearning: Language models as few shot unlearners (2024), \url{https://arxiv.org/abs/2310.07579}

\bibitem{peng2024livelearnableincontextvector}
Peng, Y., Hao, C., Yang, X., Peng, J., Hu, X., Geng, X.: Live: Learnable in-context vector for visual question answering (2024), \url{https://arxiv.org/abs/2406.13185}

\bibitem{pennington-etal-2014-glove}
Pennington, J., Socher, R., Manning, C.: {G}lo{V}e: Global vectors for word representation. In: Moschitti, A., Pang, B., Daelemans, W. (eds.) Proceedings of the 2014 Conference on Empirical Methods in Natural Language Processing ({EMNLP}). pp. 1532--1543. Association for Computational Linguistics, Doha, Qatar (Oct 2014). \doi{10.3115/v1/D14-1162}, \url{https://aclanthology.org/D14-1162/}

\bibitem{press-etal-2023-measuring}
Press, O., Zhang, M., Min, S., Schmidt, L., Smith, N., Lewis, M.: Measuring and narrowing the compositionality gap in language models. In: Bouamor, H., Pino, J., Bali, K. (eds.) Findings of the Association for Computational Linguistics: EMNLP 2023. pp. 5687--5711 (Dec 2023)

\bibitem{radford2019language}
Radford, A., Wu, J., Child, R., Luan, D., Amodei, D., Sutskever, I., et~al.: Language models are unsupervised multitask learners. OpenAI blog  \textbf{1}(8), ~9 (2019)

\bibitem{2020t5}
Raffel, C., Shazeer, N., Roberts, A., Lee, K., Narang, S., Matena, M., Zhou, Y., Li, W., Liu, P.J.: Exploring the limits of transfer learning with a unified text-to-text transformer. Journal of Machine Learning Research  \textbf{21}(140),  1--67 (2020), \url{http://jmlr.org/papers/v21/20-074.html}

\bibitem{ram-etal-2023-context}
Ram, O., Levine, Y., Dalmedigos, I., Muhlgay, D., Shashua, A., Leyton-Brown, K., Shoham, Y.: In-context retrieval-augmented language models. Transactions of the Association for Computational Linguistics  \textbf{11},  1316--1331 (2023). \doi{10.1162/tacl\_a\_00605}, \url{https://aclanthology.org/2023.tacl-1.75}

\bibitem{razeghi-etal-2022-impact}
Razeghi, Y., Logan~IV, R.L., Gardner, M., Singh, S.: Impact of pretraining term frequencies on few-shot numerical reasoning. In: Goldberg, Y., Kozareva, Z., Zhang, Y. (eds.) Findings of the Association for Computational Linguistics: EMNLP 2022. pp. 840--854 (Dec 2022)

\bibitem{Reddy2024the}
Reddy, G.: The mechanistic basis of data dependence and abrupt learning in an in-context classification task. In: The Twelfth International Conference on Learning Representations (2024), \url{https://openreview.net/forum?id=aN4Jf6Cx69}

\bibitem{rubin-etal-2022-learning}
Rubin, O., Herzig, J., Berant, J.: Learning to retrieve prompts for in-context learning. In: Carpuat, M., de~Marneffe, M.C., Meza~Ruiz, I.V. (eds.) Proceedings of the 2022 Conference of the North American Chapter of the Association for Computational Linguistics: Human Language Technologies. pp. 2655--2671 (Jul 2022)

\bibitem{schirrmeister2017}
Schirrmeister, R.T., Springenberg, J.T., Fiederer, L.D.J., Glasstetter, M., Eggensperger, K., Tangermann, M., Hutter, F., Burgard, W., Ball, T.: Deep learning with convolutional neural networks for eeg decoding and visualization. Human Brain Mapping  \textbf{38}(11),  5391--5420 (2017). \doi{https://doi.org/10.1002/hbm.23730}, \url{https://onlinelibrary.wiley.com/doi/abs/10.1002/hbm.23730}

\bibitem{serina_synonyms}
Serina, L., Putelli, L., Gerevini, A.E., Serina, I.: Synonyms, antonyms and factual knowledge in bert heads. Future Internet  \textbf{15}(7) (2023). \doi{10.3390/fi15070230}, \url{https://www.mdpi.com/1999-5903/15/7/230}

\bibitem{singh2023the}
Singh, A.K., Chan, S.C., Moskovitz, T., Grant, E., Saxe, A.M., Hill, F.: The transient nature of emergent in-context learning in transformers. In: Thirty-seventh Conference on Neural Information Processing Systems (2023), \url{https://openreview.net/forum?id=Of0GBzow8P}

\bibitem{singh2024needs}
Singh, A.K., Moskovitz, T., Hill, F., Chan, S.C., Saxe, A.M.: What needs to go right for an induction head? a mechanistic study of in-context learning circuits and their formation. arXiv preprint arXiv:2404.07129  (2024)

\bibitem{song2023decoding}
Song, Y., Liu, B., Li, X., Shi, N., Wang, Y., Gao, X.: Decoding natural images from eeg for object recognition. arXiv preprint arXiv:2308.13234  (2023)

\bibitem{Emu2}
Sun, Q., Cui, Y., Zhang, X., Zhang, F., Yu, Q., Luo, Z., Wang, Y., Rao, Y., Liu, J., Huang, T., Wang, X.: Generative multimodal models are in-context learners  (2023)

\bibitem{Suo2024VisualPS}
Suo, W., Lai, L., Sun, M., Zhang, H., Wang, P., Zhang, Y.: Visual prompt selection for in-context learning segmentation (2024), \url{https://api.semanticscholar.org/CorpusID:271213205}

\bibitem{bnci2012}
Tangermann, M., Müller, K.R., Aertsen, A., Birbaumer, N., Braun, C., Brunner, C., Leeb, R., Mehring, C., Miller, K.J., Mueller-Putz, G., Nolte, G., Pfurtscheller, G., Preissl, H., Schalk, G., Schlögl, A., Vidaurre, C., Waldert, S., Blankertz, B.: Review of the bci competition iv. Frontiers in Neuroscience  \textbf{6} (2012). \doi{10.3389/fnins.2012.00055}, \url{https://www.frontiersin.org/journals/neuroscience/articles/10.3389/fnins.2012.00055}

\bibitem{Thieen2023ProbingLL}
Thie{\ss}en, F., D'Souza, J., Stocker, M.: Probing large language models for scientific synonyms. In: SEMANTICS Workshops (2023), \url{https://api.semanticscholar.org/CorpusID:265068593}

\bibitem{52580}
Von~Oswald, J., Niklasson, E., Randazzo, E., Sacramento, J., Mordvintsev, A., Zhmoginov, A., Vladymyrov, M.: Transformers learn in-context by gradient descent. In: International Conference on Machine Learning. pp. 35151--35174. PMLR (2023)

\bibitem{Voronov2024-iv}
Voronov, A., Wolf, L., Ryabinin, M.: Mind your format: Towards consistent evaluation of in-context learning improvements. arXiv [cs.CL]  (Jan 2024)

\bibitem{wies2023learnability}
Wies, N., Levine, Y., Shashua, A.: The learnability of in-context learning. Advances in Neural Information Processing Systems  \textbf{36},  36637--36651 (2023)

\bibitem{Wu_2018_CVPR}
Wu, Z., Xiong, Y., Yu, S.X., Lin, D.: Unsupervised feature learning via non-parametric instance discrimination. In: Proceedings of the IEEE Conference on Computer Vision and Pattern Recognition (CVPR) (June 2018)

\bibitem{yang2023auto}
Yang, J., Ma, S., Wei, F.: Auto-icl: In-context learning without human supervision. arXiv preprint arXiv:2311.09263  (2023)

\bibitem{weibo2014}
Yi, W., Qiu, S., Wang, K., Qi, H., Zhang, L., Zhou, P., He, F., Ming, D.: Evaluation of eeg oscillatory patterns and cognitive process during simple and compound limb motor imagery. PLOS ONE  \textbf{9}(12),  1--19 (12 2014). \doi{10.1371/journal.pone.0114853}, \url{https://doi.org/10.1371/journal.pone.0114853}

\bibitem{NEURIPS2023_398ae57e}
Zhang, Y., Zhou, K., Liu, Z.: What makes good examples for visual in-context learning? In: Oh, A., Naumann, T., Globerson, A., Saenko, K., Hardt, M., Levine, S. (eds.) Advances in Neural Information Processing Systems. vol.~36, pp. 17773--17794. Curran Associates, Inc. (2023), \url{https://proceedings.neurips.cc/paper_files/paper/2023/file/398ae57ed4fda79d0781c65c926d667b-Paper-Conference.pdf}

\bibitem{zhou2016}
Zhou, B., Wu, X., Lv, Z., Zhang, L., Guo, X.: A fully automated trial selection method for optimization of motor imagery based brain-computer interface. PLOS ONE  \textbf{11}(9),  1--20 (09 2016). \doi{10.1371/journal.pone.0162657}, \url{https://doi.org/10.1371/journal.pone.0162657}

\bibitem{zhu2024incoroincontextlearningrobotics}
Zhu, J.Y., Cano, C.G., Bermudez, D.V., Drozdzal, M.: Incoro: In-context learning for robotics control with feedback loops (2024), \url{https://arxiv.org/abs/2402.05188}

\bibitem{zucchet2025emergencesparseattentionimpact}
Zucchet, N., d'Angelo, F., Lampinen, A.K., Chan, S.C.Y.: The emergence of sparse attention: impact of data distribution and benefits of repetition (2025), \url{https://arxiv.org/abs/2505.17863}

\end{thebibliography}

\newpage
\appendix

{\Large{\bf Supplementary Material}}\\

\section{Model details} \label{app:model_details}

\begin{figure}[!tbph]
  \centering
    \includegraphics[width=0.9\textwidth]{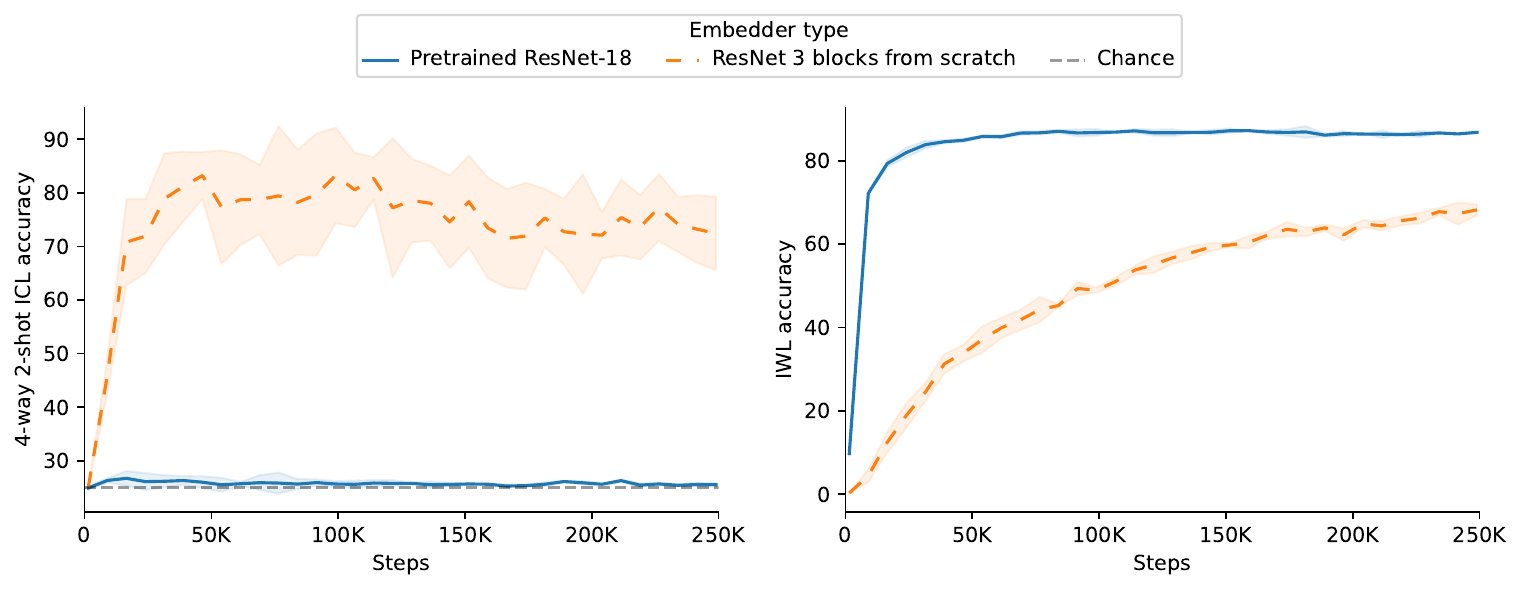}
  \caption{4-way 2-shot ICL and IWL accuracy for different versions of the image embedder. Using pretrained models makes the in-weight task easier and ICL does not emerge.}
  \label{fig:embedder}
\end{figure}

\textbf{Architecture details.} We train the GPT-2 model~\cite{radford2019language} with 12 layers and 8 heads with an embedding dimension of 64. 
We use a smaller model for the induction head analysis experiments with 3 layers, a single head, and an embedding dimension of 64. 
GPT-2 expects a sequence-like format with aligned embedding size so we transformed our image-label pairs into separate image and label tokens, using a ResNet-like embedder for images and an embedding layer for labels. We initialized the model with a truncated normal distribution, which is important for training stability.
We use a 3-block ResNet model~\cite{he2016deep} as the image embedder with output channel dimensions [64, 128, 256]. After that, a projection layer is added to match the embedding dimension of 64. We train the image embedding model and GPT model together from scratch.
We notice that the emergence of ICL is sensitive to the input image embedder architecture. We also report that pretrained embedders result in fast convergence of in-weight task and ICL failure cases, as shown in Figure \ref{fig:embedder}.

\textbf{Hyperparameters.} We trained the model for different numbers of steps varying from 250k to 2M iterations using optimizer Adam ~\cite{Kingma2014AdamAM} with betas (0.9, 0.99) and epsilon 1e-08. We use learning rate warm-up for 15K iterations with a square root decay scheduler with a maximum learning rate value of 6e-4. We find that ICL performance is enhanced with longer warm-up periods. We perform gradient clipping to value 1.0. We trained the model with a batch size of 16 on a single Nvidia RTX 3090 where 500k iterations took around 12 hours.
For all experiments, we run the approach for 3 random seeds and report averaged results.

\section{Experiment setup details} \label{app:experiment_details}

\textbf{Training details.}
The model is trained with a mix of two types of sequences with different burstiness forms: in-context sequences and standard sequences. In-context sequences can have multiple reoccurring samples from the same class as the query, whereas standard sequences have all samples selected from random classes. The high-burstiness strategy uses in-context sequences with three repetitions from the query class in context, while the instCopy strategy employs exact copies and copy-pastes the query image in the context.
All the supervised models in this work are trained with a burstiness probability of 90\%, which means 90\% in-context sequences and 10\% standard sequences. All self-supervised models use 100\% in-context sequences.

\textbf{Evaluation details.} We evaluate separately for both ICL and IWL. ICL evaluation is performed in a few-shot manner with 2-way 4-shots and 4-way 2-shots tasks. The evaluation is performed using the pretrained softmax classifier without any model update. We use label mappings 0 to $k$ with $k$ being the number of classes in the few-shot setting. We notice that the ICL results are agnostic to label mappings.
ICL is always performed on the hold-out classes not seen during training.
ICL is performed over 10K presampled sequences to ensure a fair comparison. IWL is evaluated on the hold-out samples from the training classes where the sequence has no repetitions so the model does not rely on ICL, but only on the model weights. 

\subsection{Datasets} \label{app:datasets_details}
All analysis experiments are performed on the Omniglot dataset ~\cite{omniglot}. We further show ICL results on other visual datasets Cifar-100~\cite{cifarfs}, Caltech-101~\cite{caltech}, and DTD textures~\cite{dtd} which are often used for benchmarking few-shot learning classification task. A few details about these datasets are included below:

\begin{figure}[!tbph]
    \centering
    \includegraphics[width=\textwidth]{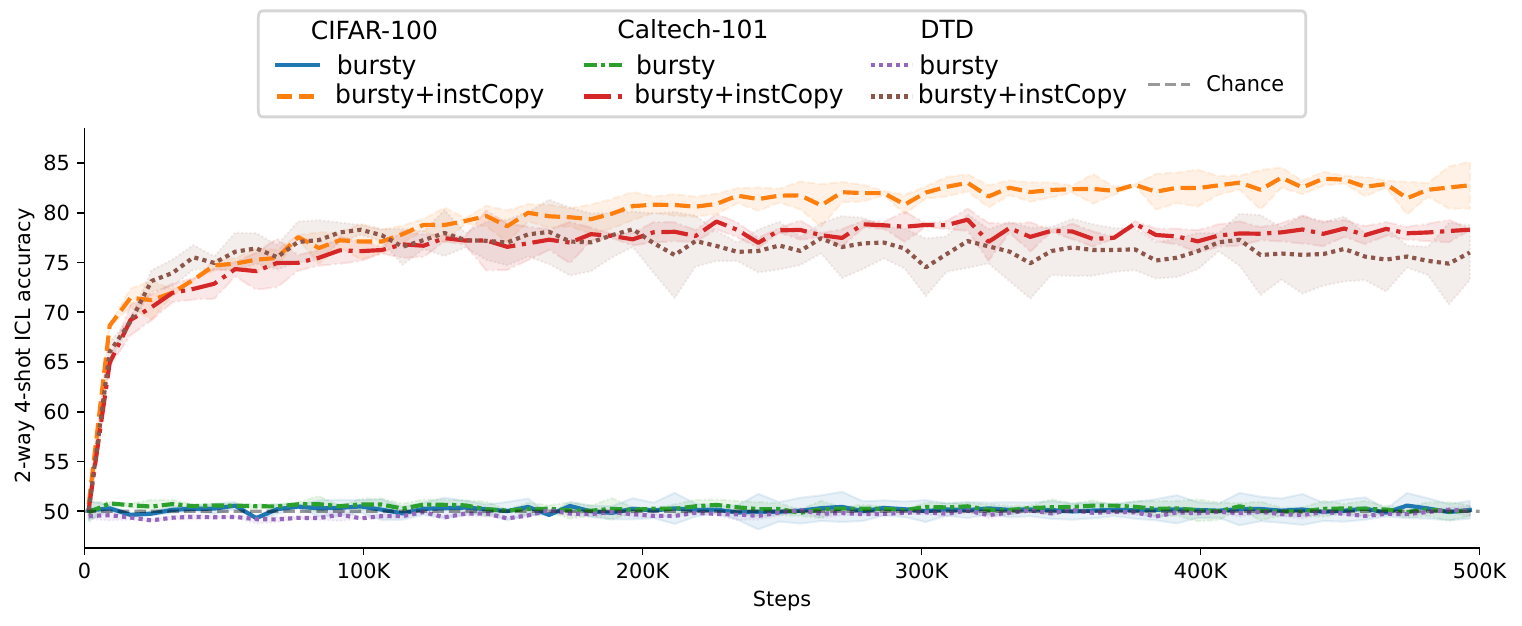}
    \caption{Including exact copies in the bursty sequences enables ICL for different visual datasets -- CIFAR-100, Caltech-101 and DTD, and results in strong and stable ICL 2-way 4-shot performance.}
    \label{fig:other_2w4s}
\end{figure}

\textbf{Omniglot} consists of 1623 handwritten characters from 50 alphabets with 20 exemplars for each character. Unless stated otherwise, we use 1600 classes as the base classes and the remaining 23 classes (sampled from the official evaluation subset with seed 42) for the ICL evaluation. We create a train-validation split as 18-2. 
During the supervised training, we apply no augmentations except for resizing to 64x64. However, self-supervised setup benefits from mild augmentations (random crop resize to 64x64 with scale (0.5, 1.5) and horizontal flip). For the self-supervised experiments in~\cref{app:iwl_task}, we used a batch size of 216 and a learning rate of 1e-3.

\textbf{CIFAR-100} is a natural dataset consisting of 60000, 32x32 colored images divided into 100 categories with 600 examples from each one. We use 80 classes for supervised training and 20 classes for the ICL evaluation as it is given by the Cifar-100FS (Few-Shot) version of the dataset. We used 10\% of the data for the validation. 
We do not apply any augmentations, but we resize the image to 64x64 for training and evaluation. 

\textbf{Caltech-101} is a natural, imbalanced dataset with 101 classes with 40-800 images per class while most classes have about 50 images and each image is roughly 300x200 pixels. We randomly select 91 classes for the supervised training and the remaining 10 classes are used for ICL evaluation. 
During training, we use random resized cropping to 64x64 with scaling from 0.5 to 1.5, horizontal flipping and random rotation of 15 degrees.

\textbf{DTD} is a texture dataset consisting of 5640 images across 47 classes with 120 images from each class with a size ranging from 300x300 to 640x640. We use 37 classes for supervised training and 10 classes for the ICL evaluation and create a train and validation split with roughly 10\% of data used for validation. 
We report better and more stable results with an image size of 128x128 and random resize with scale (0.5, 1.5).

\textbf{Results on CIFAR, Caltech and DTD.} The 4-way-2-shot ICL results are included in the main paper. Here, we show 2-way 4-shot and IWL performance for other datasets in Figure \ref{fig:other_2w4s}. We observe strong ICL performance using the combined strategy of instCopy and high burstiness while the baselines alone does not result in any ICL performance.

\section{Promoting ICL through exact repetitions} \label{app:look_up_details}

\begin{figure}[!h]
  \centering
    \begin{minipage}[t]{0.48\textwidth}
        \includegraphics[width=\textwidth]{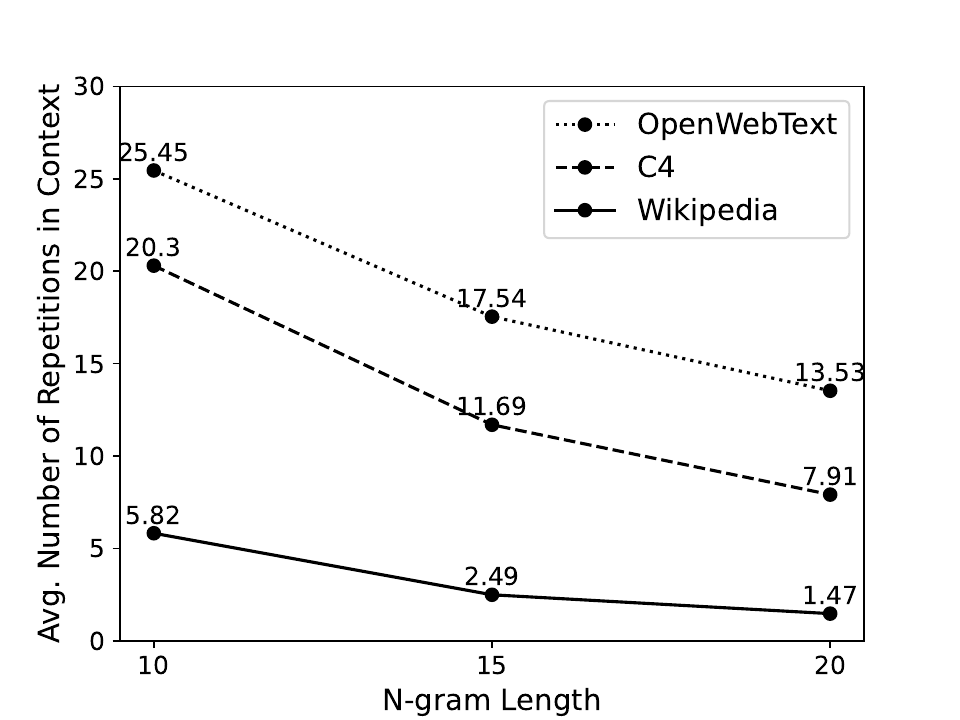}
    \end{minipage}
    \begin{minipage}[b]{0.5\textwidth}
        \justifying
        \scriptsize{Elymus, \textcolor{orange}{attended Pirithous' wedding, fought in the battle against the Lapiths and was killed by}… Eurytion, acted in an insulting manner towards Hippolyte when she was being joined in marriage to Azan or in the house of Pirithous...
        Hodites, fought against the Lapiths at Pirithous' wedding. Killed by Mopsus. Hyles, \textcolor{orange}{attended Pirithous' wedding, fought in the battle against the Lapiths and was killed by}… Imbreus, \textcolor{orange}{fought against the Lapiths at Pirithous' wedding and was killed by Dryas}… Isoples, \textcolor{blue}{killed by Heracles when he tried to steal the wine of Pholus}… Lycabas, attended Pirithous' \textcolor{dgreen}{wedding, fought against the Lapiths and fled}. Lycidas, \textcolor{orange}{fought against the Lapiths at Pirithous' wedding and was killed by Dryas}. Lycus, fought against the Lapiths at Pirithous' wedding was killed by Pirithous. Medon, attended Pirithous' \textcolor{dgreen}{wedding, fought against the Lapiths and fled}. Melanchaetes, \textcolor{blue}{killed by Heracles when he tried to steal the wine of Pholus}. Melaneus, \textcolor{dgreen}{attended Pirithous' wedding, fought against the Lapiths and fled}. Mermerus, wounded by the Lapiths at Pirithous’… }
    \end{minipage}
  \caption{Left: Repetitions of n-grams in Wikipedia\cite{wikidump}, OpenWebText\cite{Gokaslan2019OpenWeb} and C4\cite{2020t5} pretraining corpora, performed over 50 million tokens using a BPE tokenizer with a context length of 2048. We report the average number of repetitions within the 2048-token window for different n-gram lengths. The variety in the corpora's format (e.g. web, news, social media, wiki) leads to substantial differences in repetition rates.
  Right: Truncated example of a 2048-token sample from Wikipedia's pretraining corpora, highlighting exact n-gram repetitions. Different colors present different n-gram lengths (green: 10-grams, blue: 15-grams, orange: 20-grams), demonstrating both patterns in the pretraining data.} 
  \label{fig:ngrams}
\end{figure}

The presence of ICL capabilities in LLMs is well established, and many have connected it to the specific data distributional properties of natural language. Chen et al.~\cite{NEURIPS2022_77c6ccac} demonstrated that burstiness in the training sequences and skewness in data -- both inherently present in natural language -- show the emergence of ICL in simple visual tasks using the Omniglot dataset~\cite{omniglot}. Besides burstiness and skewness, we argue that natural language typically contains many exact token copies and n-grams, which, we believe, is an important factor for stable and non-transient ICL in LLMs. To test this, we performed a small analysis of three pretraining corpora datasets: Wikipedia~\cite{wikidump}, OpenWebText~\cite{Gokaslan2019OpenWeb}, and C4~\cite{2020t5}. We calculated the average number of $n$-gram repetitions within a context window of 2048 tokens, where the text was tokenized using a BPE tokenizer. In~\cref{fig:ngrams}, we report many $n$-gram repetitions, even for the longer $n$-grams of length 20, but we observe how different domains of the text (web, news, social media, wiki) influence the number of repetitions. Furthermore, we present a truncated example from Wikipedia's pretraining corpora that reports many repetitions, and indeed observe many specific patterns reappearing throughout the text. 

In this section, we present additional results and details on the effect of burstiness and exact copies for ICL emergence.

\subsection{Importance of burstiness for ICL} \label{app:burstiness}

\begin{figure}[tp!]
    \centering
    \includegraphics[width=.9\textwidth]{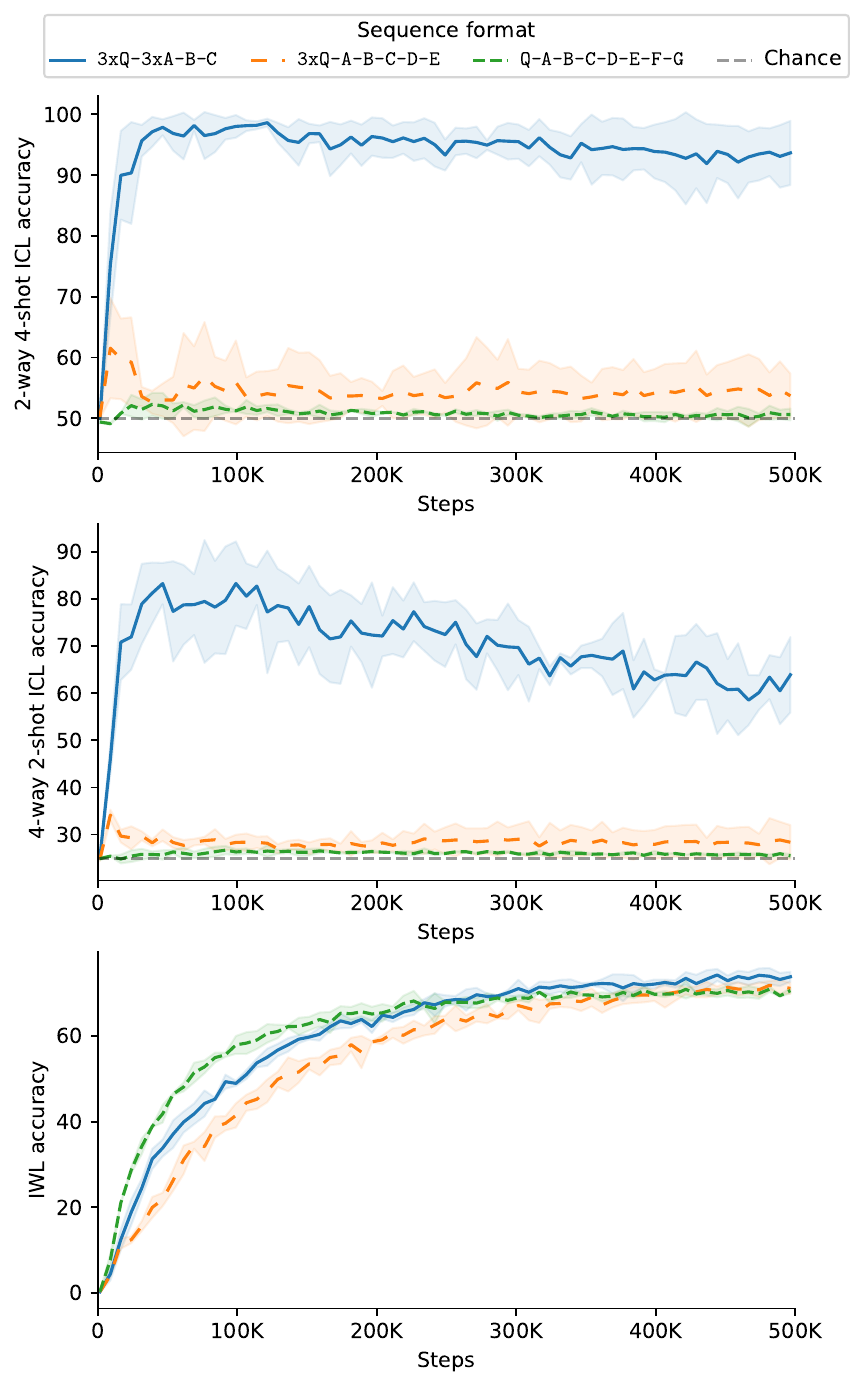}
    \caption{High burstiness improves ICL performance without affecting IWL performance. 2-way 4-shot evaluation is easier ICL setup which results in more stable and less noisy performance than 4-way 2-shot. IWL performance remains same similar with difference early in the training.}
    \label{fig:burstiness_appendix}
\end{figure}

First, we confirm previous findings about the importance of burstiness format in the training data. Here, we introduce a new notation to facilitate easier differentiation between burstiness formats. Context of the sequence (image-label pairs) as given as a sequence of C class identifiers and number of repetitions in the sequence. Thus, format \seq{3xQ-3xA-B-C} represents a high level of burstiness with three instances combing from the same class as query, three instances coming from another class, and two instances from the other two classes; this format is exactly the same as the one used in the main analysis and visualized in~\cref{fig:sequences_overview} as a bursty sequence. Furthermore, we explore two additional levels of burstiness:~\seq{3xQ-A-B-C-D-E} with three instances coming from the same class as the query, while the other five are from five different classes, and~\seq{Q-A-B-C-D-E-F-G} with all classes in the context being distinct from the query and without any repetitions. In~\cref{fig:burstiness_appendix}, we show that a larger magnitude of burstiness promotes better ICL performance, which is seen on both 4-way 2-shot and 2-way 4-shot ICL performance. We further report IWL performance where we see that bursty sequence slow down the IWL convergence early in the training, but in the end all burstiness level converge to the same value, indicating that the model in the case of bursty sequences is relying on the ICL mechanisms even during training once they have been established.

\subsection{Exact copies in burstiness further improve ICL} \label{app:instCopy}

\begin{figure}[bp!]
    \centering
    \includegraphics[scale=.45]{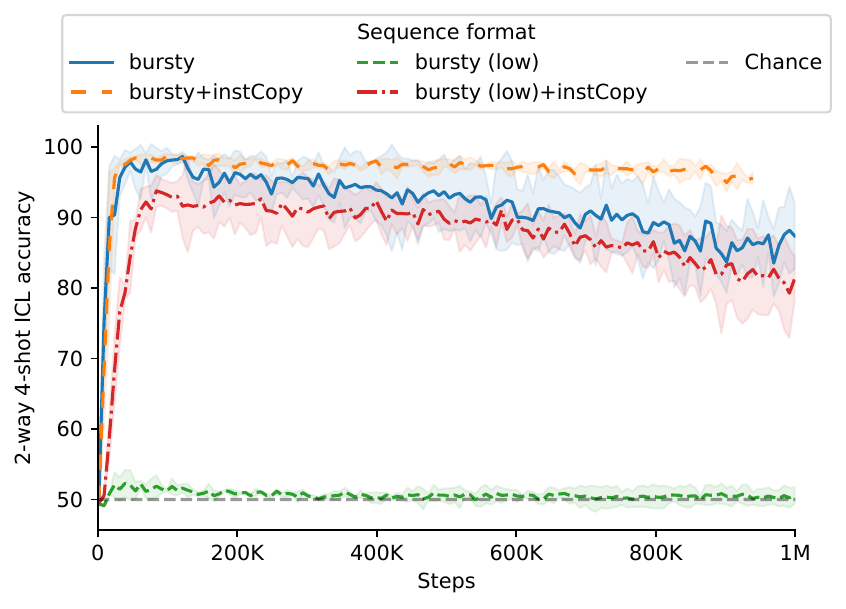}
    \caption{Inlcuding exact copies in the bursty sequences results in stronger and more stable ICL performance. Only one exact token copy is not sufficient to enable ICL, demonstrating that high burstiness is not necessary when exact copies are in the context.}
    \label{fig:repetitions_2w4s}
\end{figure}

We further demonstrate that including exact copies in the bursty sequences yields even better ICL performance, characterised by higher accuracy and reduced transiency, compared to using bursty sequences alone. Furthermore, by employing a single token copy in the context, we are now able to emerge the ICL performance, showing that, in this case, burstiness is no longer needed for ICL emergence. In~\cref{fig:repetitions_2w4s}, we present the performance for the 2-way 4-shot ICL task, where we achieve even better results than in the 4-way 2-shot case, as the ICL task is now easier.

\subsection{Deeper look into the training dynamics with exact copies} \label{app:ih_analysis}

Here, we show the formation of induction heads and ICL emergence only when using in-context sequences with exact copies in the context (bursty + instCopy). We perform an analysis using a smaller GPT-2 model with 3-layers and 1-head. We observe strong ICL performance; however, it is transient as the IWL progresses. 

\begin{figure}[h!]
    \centering
    \includegraphics[width=\textwidth]{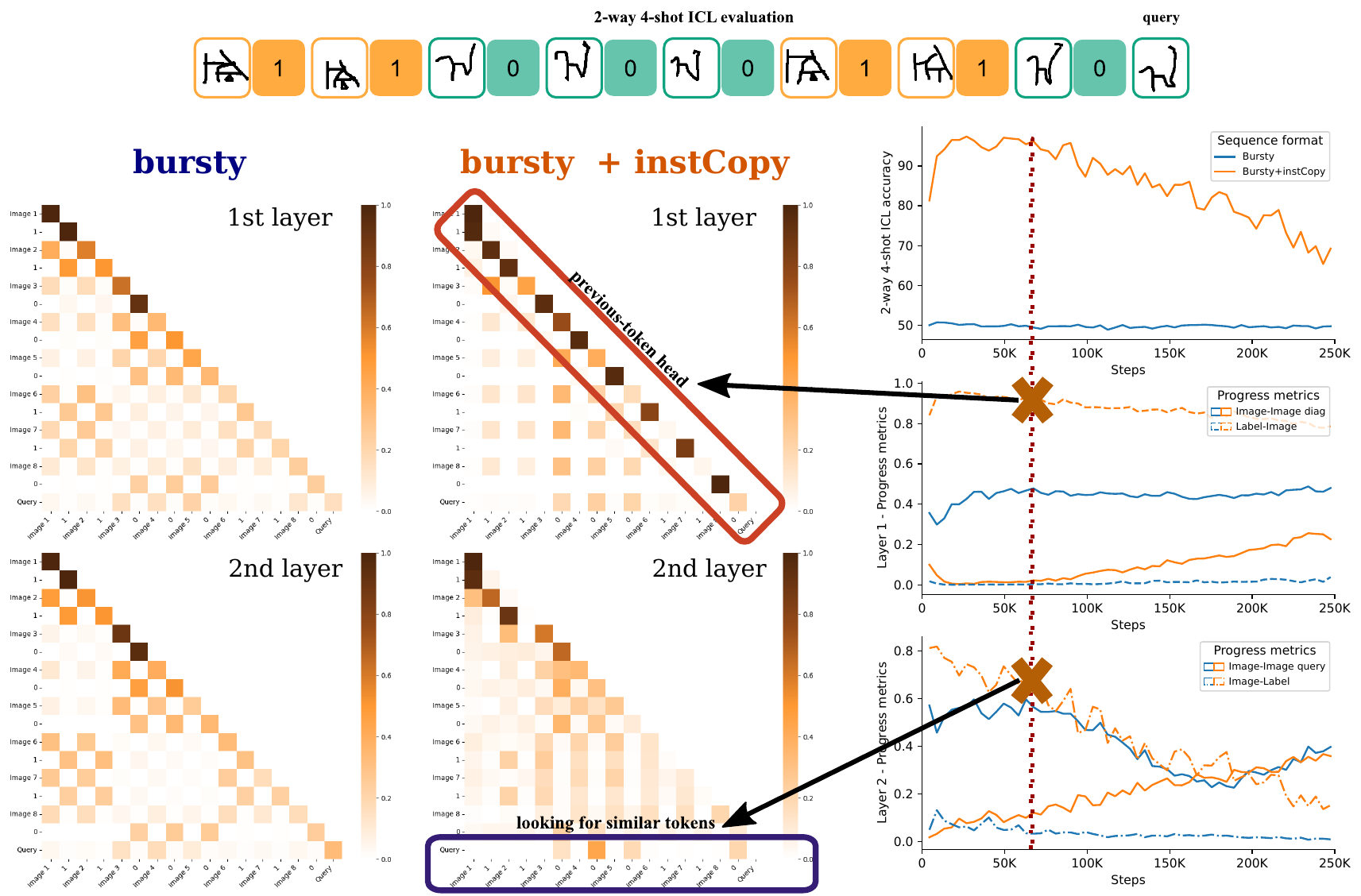}
    \caption{Induction head analysis of the GPT-2 model with 3 layers and 1 head for the models trained with and without exact copies in the bursty sequence. We observe strong ICL performance only for the model with exact copies, where the progress metrics confirm the formation of the previous-token head and the model's correct implementation of the ICL mechanisms. Interestingly, as the ICL performance becomes transient, the model starts to behave similarly to the model that has no ICL performance -- just performing modality mapping.}
    \label{fig:induction_head_measures_1s}
\end{figure}

\textbf{Progress metrics.} We study the formation of induction heads and the emergence of ICL using four progress metrics: 1) image-image diagonal, 2) label-image, 3) image-image query, and 4) image-label. 
The \textbf{image-image diagonal} measure is the average attention between all image tokens and all other image tokens on the diagonal, representing the modality mapping.
The \textbf{label-image} measures the average attention between each label token and its previous image token on the positions which are expected to be activated if the previous-token head has been formed (diagonal moved by one).
The \textbf{image-image query} measures the average attention between the query image and other images from the same class, representing the similarity matching function instead of the knowledge aggregation.
Finally, the \textbf{image-label} measures the average attention between the query image and the correct label token positions, which represents the final step in the induction head -- retrieval of the correct label.

In~\cref{fig:induction_head_measures_1s}, we show the progress measure calculated on the ICL sequences throughout the training, and we visualize the QK attention space of one sequence at 65k iterations when a model with exact copies in the sequence had strong ICL performance. In contrast, the one with just burstiness had no ICL performance. We observe different patterns and trends in the progress measures for the two models.

In Layer 1, we observe a high label-image progress metric for the model with exact copies, indicating that the label tokens now strongly attend to the previous image in the sequence, which confirms the formation of the previous-token head. Interestingly, for the model with just burstiness, we observe no such formation. Instead, the model learns the modality mapping -- image tokens attend to other image token positions in the sequence. 

In Layer 2, we initially observe a high image-label progress metric for the model with exact copies, but these values become transient throughout training. At the same time, ICL performance follows the same trend - initially high and strong but then becomes transient. This measure directly reflects the ICL ability in a model, and we indeed observe high values only for the model that has ICL emergence. Furthermore, we observe that the model without ICL performs similarity matching and modality attendance, as evidenced by high image-to-image query values. Interestingly, we further observe an increase in the same measure as the model's ICL performance becomes transient.

\section{Role of IWL difficulty for ICL emergence} \label{app:iwl_task}

As motivated earlier, we showed that the emergence of ICL can be slowed down or completely nonexistent if the IWL is overly simplistic. In such cases, the model prioritizes the learning of IWL, completely ignoring bursty in-context sequences and ICL mechanisms. In~\cref{sec:iwl_other} we show how indeed ICL emerge once the IWL task is made harder by increasing the number of classes or introducing label noise through label swapping. Here, we present two additional methods for making the IWL more challenging: adjusting the number of samples used for training through skewness and switching to a significantly harder training objective, namely instance discrimination.

\paragraph{Number of classes} \label{app:classes}
As previously mentioned, we observe that increasing the number of classes monotonically improves ICL performance. This finding follows similar insights in prior work~\cite{NEURIPS2022_77c6ccac,Reddy2024the}. However, these works explain the improved ICL capabilities by the large number of rarely occurring classes. We interpret it as just one way among many to make the IWL task harder.

\begin{figure}[!h]
    \centering
    \begin{minipage}[b]{0.45\textwidth}
        \includegraphics[width=\textwidth]{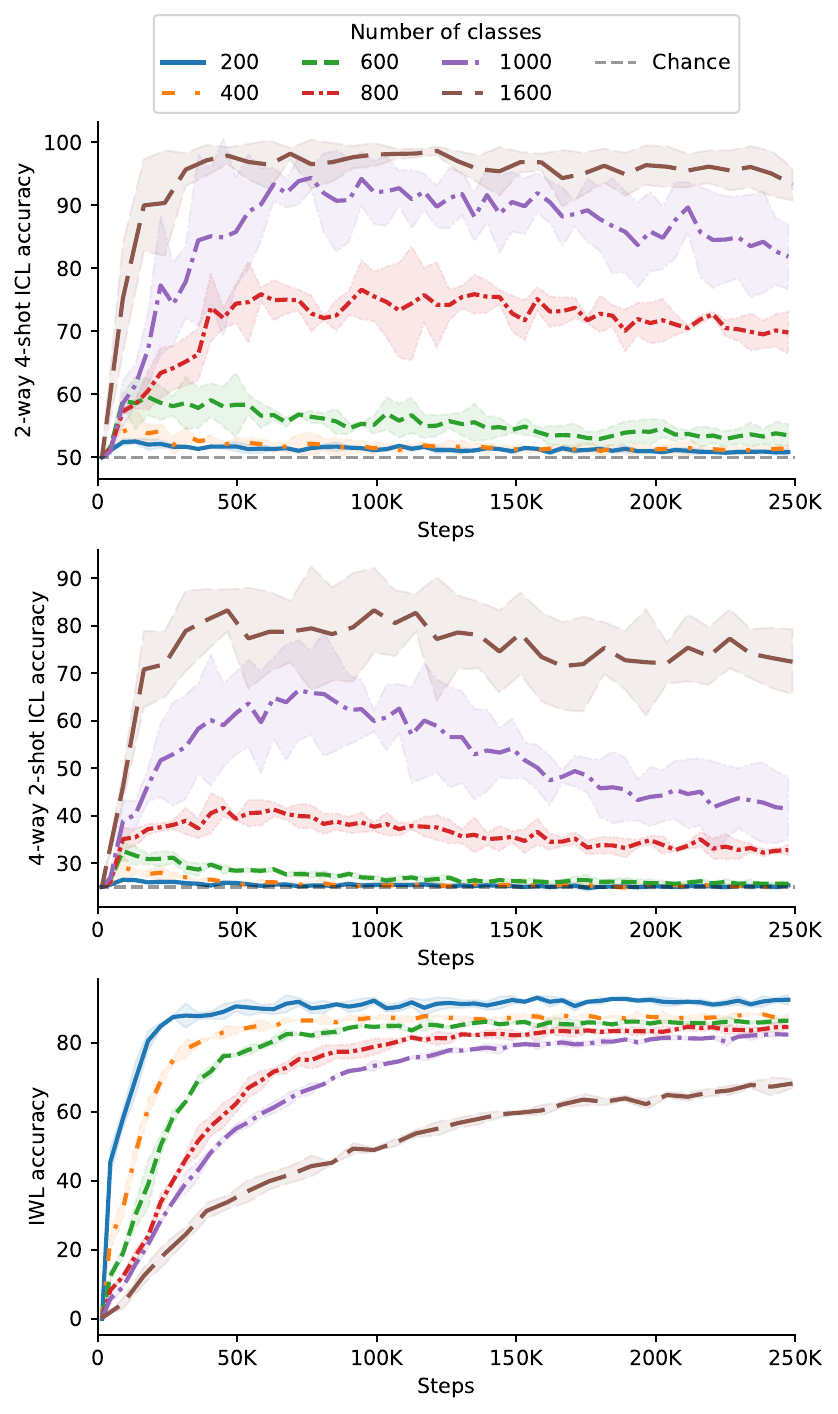}
        \caption{Increasing number of classes makes the IWL task harder, which is seen by slower IWL convergence. At the same time, ICL is being enabled, showing how certain IWL difficulty can drive ICL emergence.}
        \label{fig:num_classes_all3}
    \end{minipage}
    \hfill
    \begin{minipage}[b]{0.45\textwidth}
        \includegraphics[width=\textwidth]{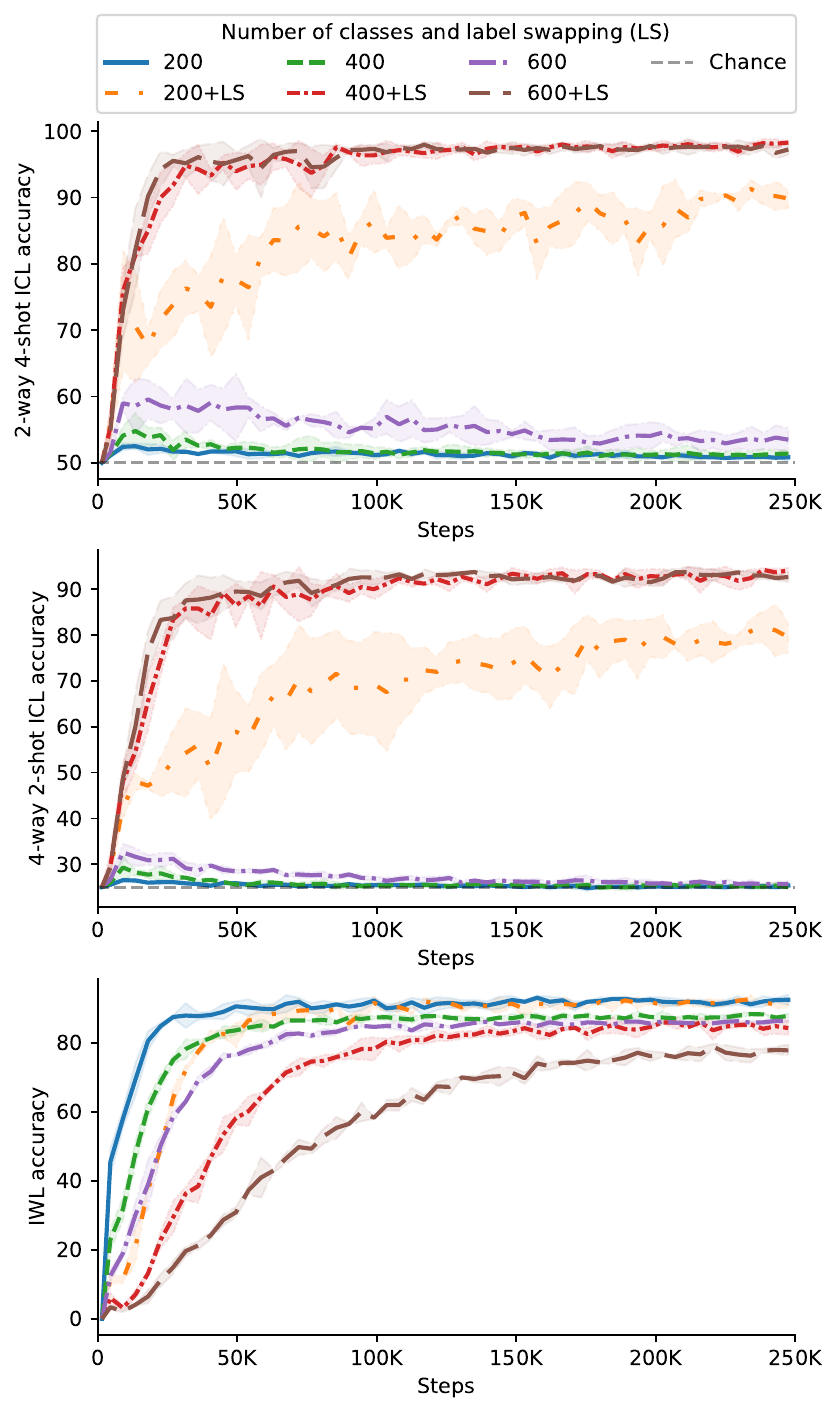}
        \caption{Adding label noise through label swapping enables ICL as the IWL task gets harder. Label swapping places more emphasis on the bursty in-context sequences, allowing the model ICL.}
        \label{fig:app_noise}
    \end{minipage}
\end{figure}

To simulate more challenging IWL scenarios, we gradually increase the number of training classes from 200 to 1600. We observe poor and unstable ICL performance until 600 classes. ICL performance significantly improves for 1000 classes; however, it remains unstable, with occasional ICL failure cases. However, we observe strong and stable ICL performance for a high number of classes as shown in \cref{fig:num_classes_all3} where we report the strong ICL accuracies for the easier (2-way 4-shot) and harder (4-way 2-shot) setup. Impaired IWL accuracy for a higher number of classes shows the IWL task is now more difficult.

\paragraph{Label noise through label swapping} \label{app:label_noise}

We further hypothesize that the model ignores the context of the bursty sequences when the IWL task is overly simplistic. We test this hypothesis by introducing label swapping to the sequences. We trained the models on bursty sequences without exact copies, using a reduced number of classes, specifically 200, 400, and 600, where no ICL was observed without noise, as the IWL task was overly simple. We now perform random label swapping, where labels of the query items are randomly assigned to another training class in 20\% of all sequences. For the case of a bursty sequence, we modify the label mappings to include all repetitions in sequences belonging to the query class.

In~\cref{fig:app_noise} we observe clear improvement in ICL performance for the models trained with label swapping. Further, comparing IWL convergences, we can see that the IWL is progressing slower when label swapping is performed, indicating that the IWL task is indeed now harder. Label swapping on standard sequences introduces noise during training, confusing the model about the correct label mappings.  At the same time, confusion over bursty sequences promotes the learning of ICL mechanisms because the model needs to condition to the context to identify the swapped label.  We demonstrate that this simple intervention improves ICL performance, even in cases where ICL was not previously emerging, such as with 200 classes. This again confirms our hypothesis that learning the ICL mechanisms can be promoted by making the IWL more challenging and enforcing the model to attend to the context.  

\begin{figure}[!h]
    \centering
    \begin{minipage}[b]{0.45\textwidth}
        \includegraphics[width=\textwidth]{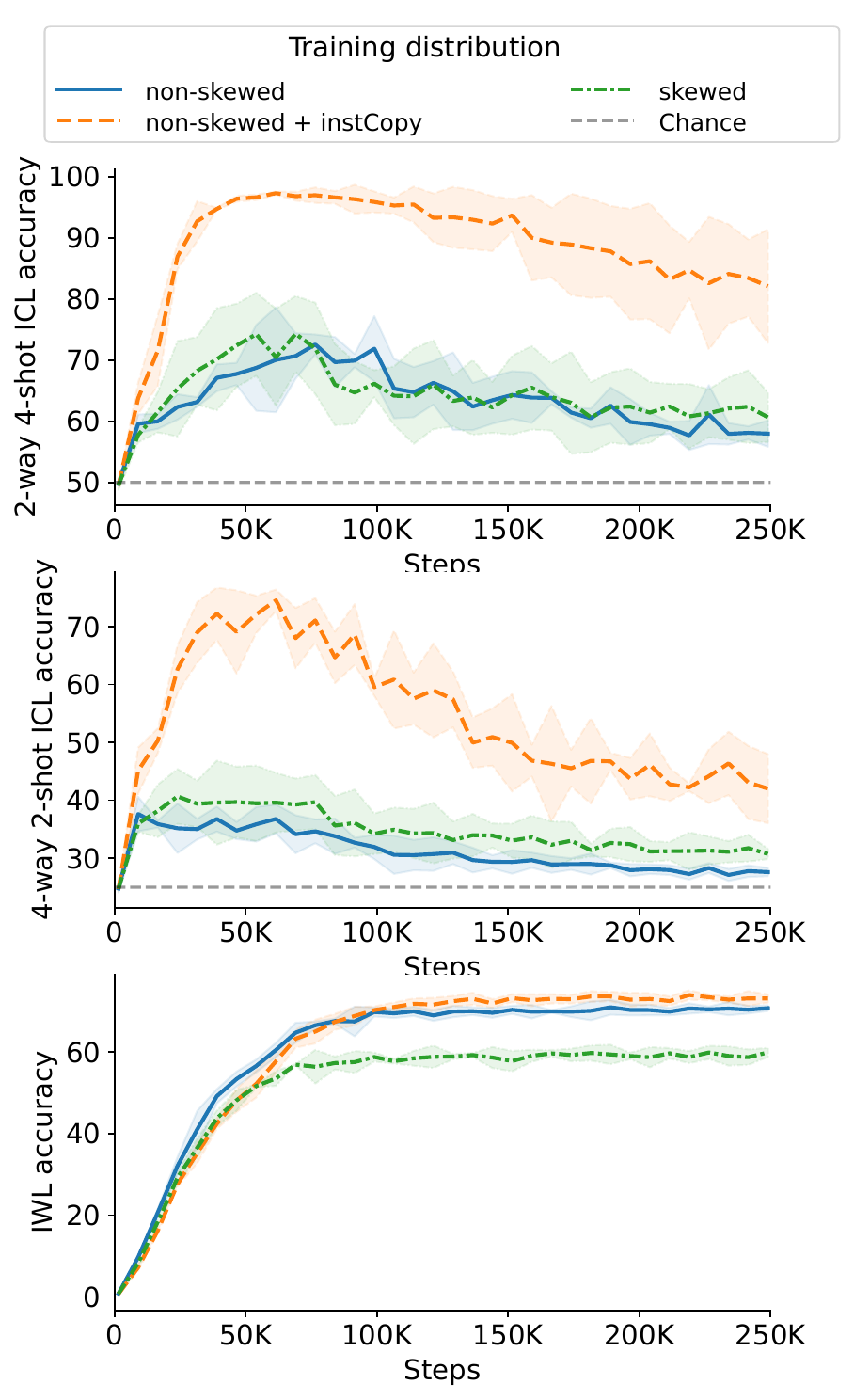}
        \caption{Skewed data enables better ICL emergence than balanced data, but using exact copies with balanced data ensures better ICL emergence. }
        \label{fig:num_classes_all3_zipf}
    \end{minipage}
    \hfill
    \begin{minipage}[b]{0.45\textwidth}
        \includegraphics[width=\textwidth]{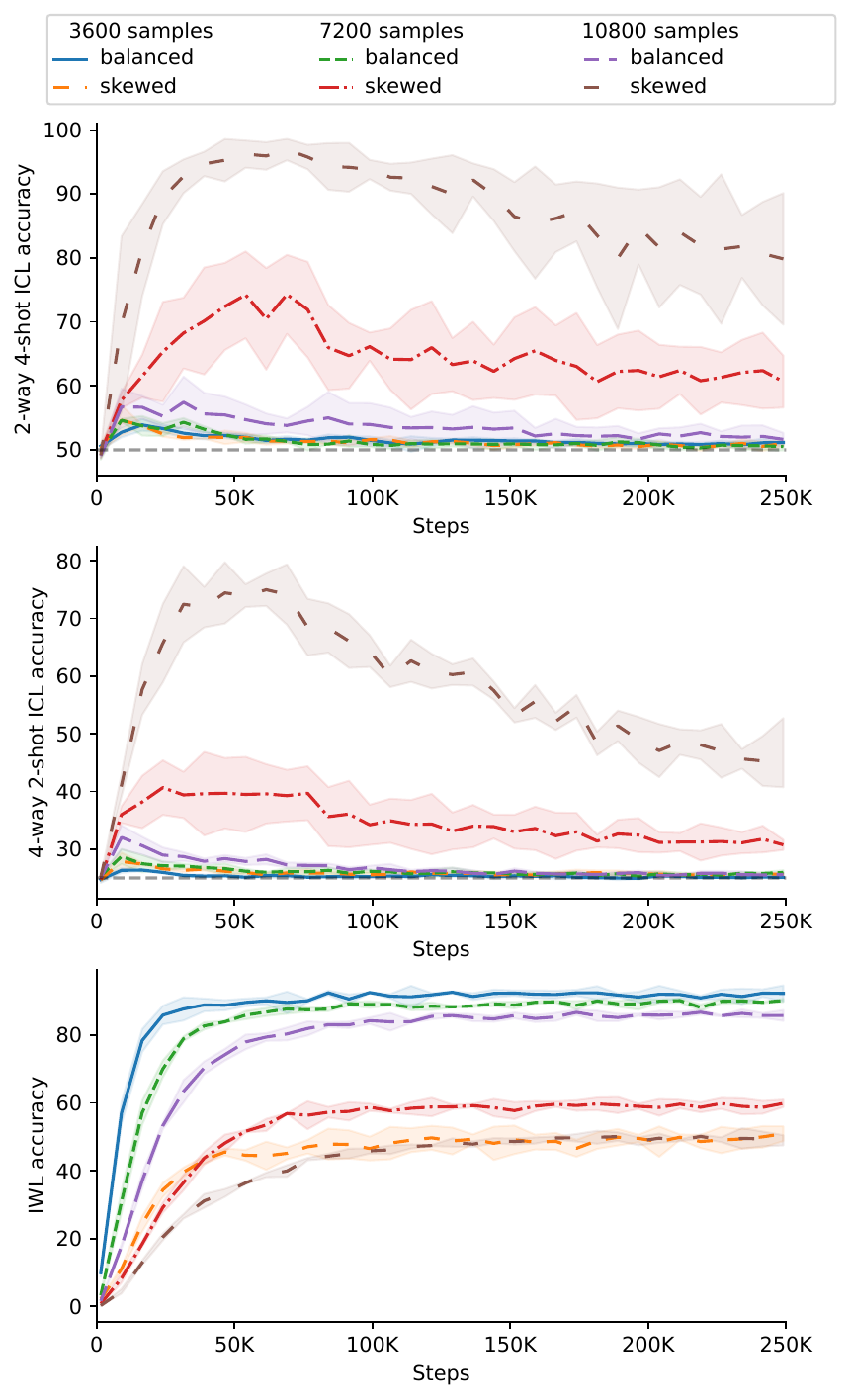}
        \caption{Skewed data distribution results in better ICL performance than balanced data over 3 different sample sizes.}
        \label{fig:samples_diff_size}
    \end{minipage}
\end{figure}

\paragraph{Skewed distribution} \label{app:skewness}
Keeping the total number of samples the same, we compare the ICL performance of a model trained with balanced and imbalanced (Zipfian) distributions. We compare skewness across different sample sizes used for training. We use three balanced datasets: 3600 samples across 200 classes, 7200 samples across 400 classes, and 10800 samples across 600 classes. These are compared to imbalanced datasets: 3598 samples across 463 classes, 7200 samples across 992 classes, and 10798 samples across 1551 classes, using a Zipfian distribution with a coefficient of 1.0.  We observe improved ICL performance with skewed distribution, as illustrated in~\cref{fig:num_classes_all3_zipf} and ~\cref{fig:samples_diff_size}. This experiment confirms prior work~\cite{NEURIPS2022_77c6ccac}, which also demonstrates improved ICL with the increased long-tail distribution. However, including exact token copies in the bursty sequences of the balanced data yields better performance, indicating that exact copies in the sequence are a strong driver for ICL performance.

\paragraph{Instance discrimination task}\label{sec:selfsup} 
Using the copy-based strategy with repetitions in the sequence, we devise a more complex IWL task by moving from supervised to self-supervised learning. We design a task based on instance discrimination~\cite{Wu_2018_CVPR}, where the model is trained to classify each sample as its corresponding class. \footnote{This would mean that in the case of the Omniglot instance discrimination setting with 1600 classes and 18 exemplars per class, the total number of classes would be 28800.} 
We train the baseline model in the supervised high-burstiness setting with 3,600 samples from 200 classes, where the ICL does not emerge. However, when we train with the instance discrimination objective using the same number of samples, we obtain very strong and stable ICL performance, which is a result of the hard IWL task and model now promoting the learning of ICL. 

\begin{figure}[h!]
    \centering
    \includegraphics[width=0.5\linewidth]{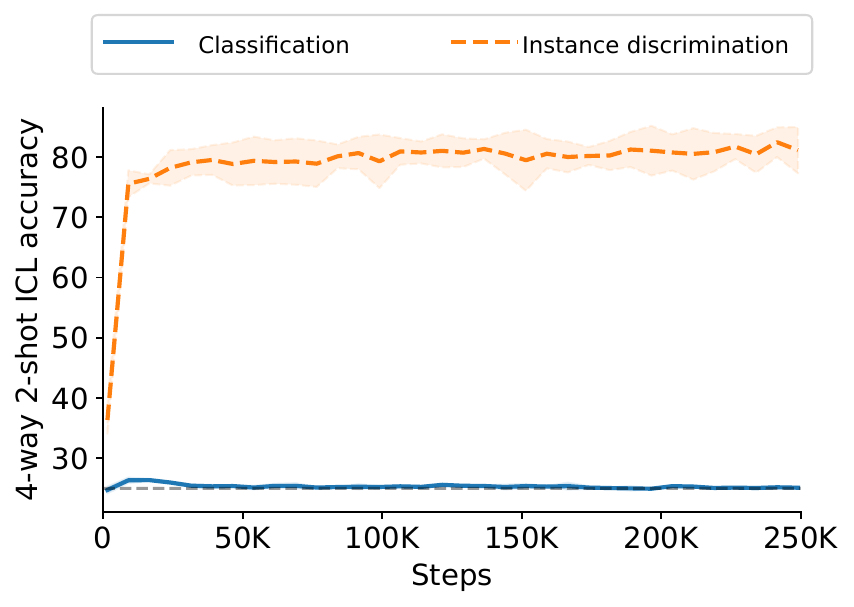}
    \caption{Moving the IWL task from multi-class classification to instance discrimination encourages the ICL learning as the IWL task is now much harder.}
    \label{fig:app_selfsup}
\end{figure}

\section{Application to EEG Classification} \label{app:eeg}

We extend the insights of our analysis to the real-life application of EEG BCI classification, where one of the main challenges is the large variability between subjects and different setups (datasets). We construct a GPT-based model trained from scratch on five widely used brain motor imagery datasets and show that our model employs cross-dataset ICL generalization.

\subsection{Experimental setup.} \label{app:eeg_experimental}

\paragraph{Architectural details.}
To enable ICL capabilities, we frame the problem as a next token prediction task following the setup used for our initial analysis with slight architectural changes for EEG data. To this end, we extend standard EEG encoders D4~\cite{schirrmeister2017} with GPT-2~\cite{radford2019language} and train the resulting overall architecture end-to-end and from scratch. D4 ~\cite{schirrmeister2017} is a simple yet effective convolutional network that produces one token for each EEG input point. Our GPT-2 backbone consists of 12 layers, 8 heads and 128-dimension hidden states. We further employ QK normalization when using the HGD~\cite{schirrmeister2017} dataset as the novel dataset, and in all experiments, we use CutCat augmentation \cite{cutcat} for improved convergence and robustness. We trained the models using AdamW~\cite{Loshchilov2017DecoupledWD} optimizer with learning rate of 5e-4, betas (0.9, 0.999), epsilon 1e-8 and weight decay 1e-2. We use learning rate warm-up for 15K iterations with a square root decay scheduler with a maximum learning rate value of 5e-4. We perform gradient clipping to value 1.0. We trained the model with a batch size of 16 on a single Nvidia RTX 4090, where we train for different amounts of iterations specific to the hold-out dataset used for ICL. Specifically, we use 125k, 150k and 250k iterations for BNCI, HGD and Zhou datasets as novel datasets, respectively. 

\paragraph{Datasets.}
We evaluate our approach on five widely-used motor imagery datasets~\cite{bnci2012, schirrmeister2017, physionet2000, weibo2014, zhou2016}. To assess generalization capabilities, we employ a leave-one-out strategy where we train on four datasets and evaluate on the fifth, permuting through all possible combinations. All datasets are preprocessed to a common format with 200 Hz sampling rate and 3 second trial windows, spanning from 0.5 seconds pre-stimulus to 2.5 seconds post-stimulus onset. Signals were bandpass filtered between 0.1 Hz and 60.0 Hz, followed by exponential moving standardization. Due to architectural constraints in the encoders requiring consistent channel configurations, we retain only the nine channels shared across all datasets (C3, C4, CP3, CP4, CPz, Cz, FC3, FC4, FCz).
The datasets differ in their class structures and sizes: BNCI~\cite{bnci2012} contains four classes (left hand, right hand, tongue, feet),~\cite{schirrmeister2017} High Gamma Dataset (HGD) includes four classes (left hand, right hand, feet, rest), PhysionetMI~\cite{physionet2000} comprises five classes (feet, hands, left hand, rest, right hand) with high class imbalance between hands and other classes, Weibo~\cite{weibo2014} features seven classes (feet, hands, left hand, right hand, left hand right foot and right hand left foot), and Zhou~\cite{zhou2016} contains three classes (feet, left hand, right hand).

We create train and validation splits by splitting the subjects in the training datasets 80-20 for training and cross-subject testing. We further split the training subjects 80-20 by trials to create training and validation splits. 

\paragraph{Training sequence construction.} \label{app:eeg_train_seq}
For an EEG trial $i$, the EEG encoder produces a single token representation $t_i$. The corresponding class label $y_i$ is embedded using a linear embedding layer.  We obtain a training sequence by interleaving $t$ and $y$: $[t_1, y_1, \dots, t_i, y_i, t_q, y_q]$, where the sequence up to the $i$-th index is used as context. The model's objective is to predict the label $y_q$ for the input EEG signal $t_q$ as shown in ~\cref{fig:gpt_overview} for the original analysis, but the same principles transfer. Unlike standard GPT training, we only compute the loss for the last predicted token $y_q$, the label corresponding to the EEG token $t_q$, referred to as the query.

The training sequences are constructed to encourage ICL, following the insights on exact token copy repetitions in context from~\cref{sec:how_to_enable_icl}. We use  90\% bursty sequences containing 3 shots for 3 distinct classes and 10\% of standard sequences without any repetitions in the context. 
For the bursty sequences, we further employ exact copies (instCopy), where we found that for BNCI~\cite{bnci2012} and Zhou~\cite{zhou2016} as novel datasets, applying exact copies in 90\% of bursty sequences works best. For HGD~\cite{schirrmeister2017}, we always employ exact copies in bursty sequences. Furthermore, we introduce label noise through label swapping in both types of sequences, as this makes the IWL task more challenging and further enhances ICL performance. We use label swapping in 10\% of sequences for BNCI and Zhou, while in 15\% of sequences for HGD dataset. We found these combinations to work the best, providing a good balance between learning the IWL task and forming needed ICL mechanisms.

\paragraph{Evaluation Details.}
For IWL evaluation, we sample sequences without class repetition in the context. ICL and generalization to unseen datasets are assessed using 3-way 3-shot classification on unseen datasets, where context samples (few-shot samples) and query samples (both unseen) are drawn from different subjects to simulate clinical deployment conditions. We sampled all possible combinations of the subject pairs for evaluation and reported average results over them. We used presampled combinations of the ICL sequences to ensure a fair comparison between runs. For all experiments, we report the average results from three runs with different seeds.

\end{document}